\documentclass[letterpaper, 10 pt, journal, twoside]{IEEEtran}

\IEEEoverridecommandlockouts                              

\usepackage{graphics} 
\usepackage{epsfig} 
\usepackage{amsmath} 
\usepackage{amssymb}  
\usepackage{algorithm}
\usepackage{algpseudocode}
\usepackage{mathtools}
\usepackage{graphicx,subfigure}
\usepackage[usenames,dvipsnames]{color}
\usepackage[nospace]{cite}
\usepackage{hyperref}
\usepackage{tikz}

\newcommand\copyrighttext{%
	\footnotesize \textcopyright 2018 IEEE. Personal use of this material is permitted.
	Permission from IEEE must be obtained for all other uses, in any current or future
	media, including reprinting/republishing this material for advertising or promotional
	purposes, creating new collective works, for resale or redistribution to servers or
	lists, or reuse of any copyrighted component of this work in other works.
	DOI: \href{https://doi.org/10.1109/LRA.2018.2856915}{10.1109/LRA.2018.2856915}}
\newcommand\copyrightnotice{%
	\begin{tikzpicture}[remember picture,overlay]
	\node[anchor=south,yshift=10pt] at (current page.south) {\fbox{\parbox{\dimexpr\textwidth-\fboxsep-\fboxrule\relax}{\copyrighttext}}};
	\end{tikzpicture}%
}

\graphicspath {{figure/}}

\newcommand{\argmax}{\operatornamewithlimits{argmax}}

\author{Jung-Su Ha, Hyeok-Joo Chae, and Han-Lim Choi%
	\thanks{Manuscript received: February, 24, 2018; Revised May, 29, 2018; Accepted June, 24, 2018.}
	\thanks{This paper was recommended for publication by Editor Nancy Amato upon evaluation of the Associate Editor and Reviewers' comments. This work was supported by the Agency for Defense Development under contract UD150047JD.}
	\thanks{J.-S. Ha, H.-J. Chae and H.-L. Choi are with the Department of Aerospace Engineering, KAIST,  Korea {\tt\small \{wjdtn1404, chhj0901, hanlimc\}@kaist.ac.kr}}%
	\thanks{Digital Object Identifier (DOI): see top of this page.} }

\begin{document}

\title{Approximate Inference-based Motion Planning by Learning and Exploiting Low-Dimensional Latent Variable Models}
\maketitle
\copyrightnotice

\markboth{IEEE Robotics and Automation Letters. Preprint Version. June, 2018} {Ha \MakeLowercase{\textit{et al.}}: Approximate Inference-based Motion Planning by Learning and Exploiting Low-Dim LVMs}

\begin{abstract}
This work presents an efficient framework to generate a motion plan of a robot with high degrees of freedom (e.g., a humanoid robot).
High-dimensionality of the robot configuration space often leads to difficulties in utilizing the widely-used motion planning algorithms, since the volume of the decision space increases exponentially with the number of dimensions.
To handle complications arising from the large decision space, and to solve a corresponding motion planning problem efficiently, two key concepts are adopted in this work:
First, the Gaussian process latent variable model (GP-LVM) is utilized for low-dimensional representation of the original configuration space.
Second, an approximate inference algorithm is used, exploiting through the duality between control and estimation, to explore the decision space and to compute a high-quality motion trajectory of the robot.
Utilizing the GP-LVM and the duality between control and estimation, we construct a fully probabilistic generative model with which a high-dimensional motion planning problem is transformed into a tractable inference problem.
Finally, we compute the motion trajectory via an approximate inference algorithm based on a variant of the particle filter.
\end{abstract}

\section{Introduction}
\IEEEPARstart{F}{or} robotic motion planning, a trajectory is designed for robot states through a complex configuration space from an initial state to perform a given task.
The planning problem is formulated as an optimal control (OC) problem considering the robot dynamics for a feasible motion trajectory and the cost function for the task.
The optimal trajectory is reconstructed from the optimal cost-to-go function (also called the value function), which is obtained by solving the Bellman equation through dynamic programming procedures.
Though these approaches guarantee the global optimality of the solution, they are not scalable with the dimensionality of the decision space because of the curse of dimensionality: the size of the decision space increases exponentially with the number of dimensions.
Sampling-based algorithms such as RRT*~\cite{karaman2011sampling} or FMT*~\cite{janson2015fast}, and their variants, are also widely used in the motion planning literature.
These algorithms simultaneously construct and extend the approximate state space (represented in a graph or a tree structure), and then update the approximate solution.
They are applicable to medium-sized problems but it is almost impossible to extend the graph into extremely high-dimensional space and obtain the solution without limiting the sampling space.
Another option is a trajectory optimization based on iterative local optimization~\cite{todorov2005generalized,toussaint2009robot}.
These approaches approximate the problem around the current solution using the first or second order Taylor expansion and update the solution iteratively.
The solution from these approaches only guarantees local optimality from its nature.
For high-dimensional problems, it is often required to give a valuable initial guess for the optimization, which is then painstakingly handcrafted by a user/designer, because the problem may have many poor local optima.
In addition, the local approximation for all dimensions can cause the solution procedure to diverge, which enforces heuristic techniques to be used.

Given this limitations, one research question that may arise is whether or not the decision space we need to search should be this large.
It may be that the dimension of the decision space really needed to be considered is small, if there is some prior structures in decisions~\cite{vernaza2012learning}.
Think about a motion planning process of our own body; when we have decided on which type of motion we are going to make, the sequence of our poses and the configurations we really need to consider may be limited. For example, we would not consider walking on our hands when the mission is to get to a certain goal position from a start position. 
That is, it is reasonable to assume that, although a robot has very high degrees of freedom, the valuable configurations take up a very small portion in the original space, and that they form some sort of (low-dimensional) manifolds.
This is the spirit behind the latent variable model.
There have been many studies aimed at finding the embedding of manifolds in high-dimensional space within a low-dimensional latent space.
Specifically, the Gaussian process latent variable model (GP-LVM) assumes that there exists a probabilistic model for mapping from low-dimensional latent space to high-dimensional observation space, and then finds the mapping and the corresponding latent space~\cite{lawrence2005probabilistic}.
Moreover, the Gaussian process dynamical model (GPDM) extends the GP-LVM such that dynamics in the observation space can be represented by that of the latent space~\cite{wang2008gaussian}.
Human motions in 50+dimensional configuration space have successfully been embedded into 2-3 dimensional latent space by GPDM, and the model has been utilized to generate new motions of a human character~\cite{wang2008gaussian,urtasun2008topologically,levine2012continuous} and for 3-D tracking of people~\cite{urtasun20063d}.
Recently, with the successes in deep learning, there have been many attempts utilizing deep neural network to construct the latent variable model;
one of the most popular algorithms is the variational autoencoder (VAE)~\cite{kingma2014auto}.
The algorithm has also been applied to robotics applications:
\cite{chen2016dynamic} exploited the idea of VAEs to embed dynamic movement primitives into the latent space.
\cite{ichter2017learning} used the conditional VAEs to learn a non-uniform sampling methodology of a sampling-based motion planning algorithm.
\cite{finn2017deep} constructed a latent dynamical system from videos and utilized the learned model to compute an optimal control policy in a model-predictive control scheme.
In \cite{ha2018adaptive,ha2018adaptive2}, we presented a variational inference method to train latent models parameterized by neural networks and utilized the learned model for planning in a similar way with this work.

One theoretical concept this work extensively takes advantage is the duality between the optimal control and estimation~\cite{todorov2008general,rawlik2012stochastic,theodorou2010generalized,todorov2011finding,hamalainen2015online,mukadam2017continuous}.
The idea is that, if we consider an artificial binary observation whose emission probability is given by the exponential of a negative cost, an OC problem can be reformulated as an equivalent inference problem.
In this case, the objective is to find the trajectory or control policy that maximizes the likelihood of the observations along the trajectory.
To address the transformed inference problem, several approximate inference techniques were utilized.
If transitions between time steps are made approximately Gaussian~\cite{toussaint2009robot} with the current control policy, the resulting algorithm becomes (locally) equivalent to the iterative linear quadratic Gaussian method~\cite{todorov2005generalized}.
In path integral control approaches, particles propagated by the current approximate control policy~\cite{theodorou2010generalized}, or the control policy induced by a higher-level path planner~\cite{ha2016topology}, are used to approximate the resulting distribution of the inference.
The user-designed probability model can also be utilized as a priori to solve the transformed inference problem~\cite{hamalainen2015online}.
Especially in~\cite{mukadam2017continuous}, the planning problem is converted into the inference problem for a factored graph of a continuous motion trajectory represented by Gaussian process, and an incremental inference method, Bayes Trees, is utilized to efficiently perform replanning procedures.
Finding a valuable proposal distribution is essential to the efficient inference method; just like finding a valuable search-space is essential for the planning problem.


\subsection{Overview and Contribution}
\begin{figure}[t]
	\centering
	\subfigure[]{
		\includegraphics*[width=.15\columnwidth]{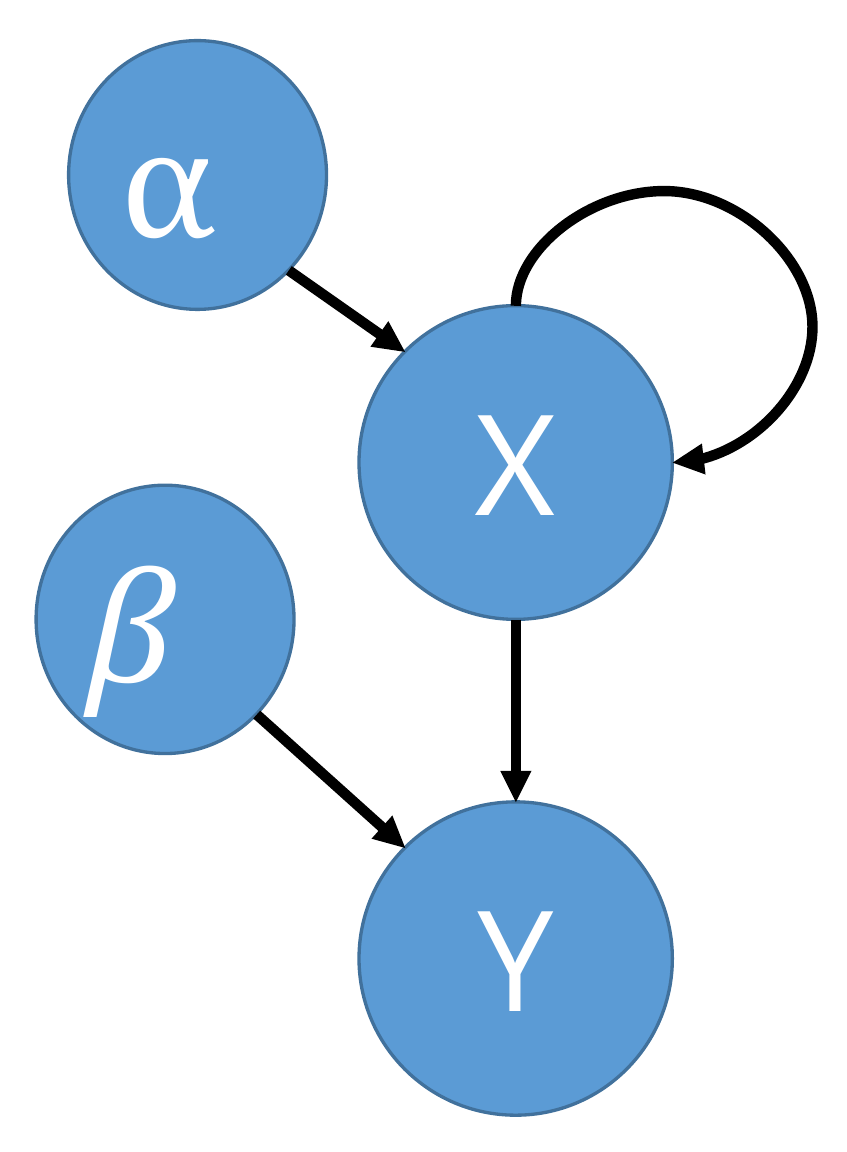}\label{fig:GPDM}}
	\subfigure[]{
		\includegraphics*[width=.11\columnwidth, viewport =210 180 390 480]{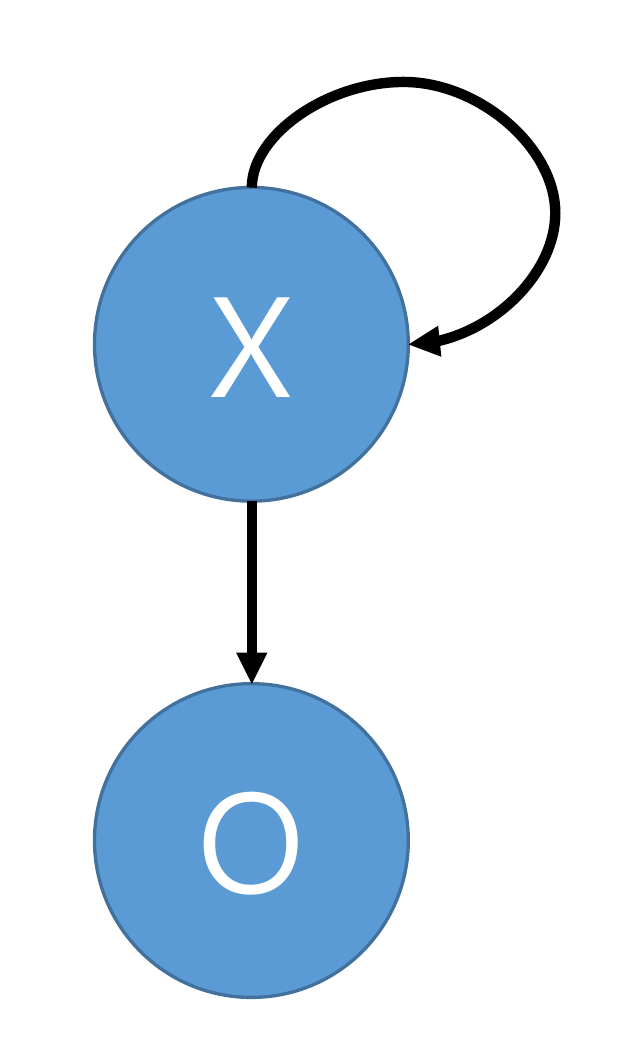}\label{fig:dual}}
	\subfigure[]{
		\includegraphics*[width=.18\columnwidth, viewport =500 50 835 480]{graph.pdf}\label{fig:motion_planning0}}
	\subfigure[]{
		\includegraphics*[width=.36\columnwidth]{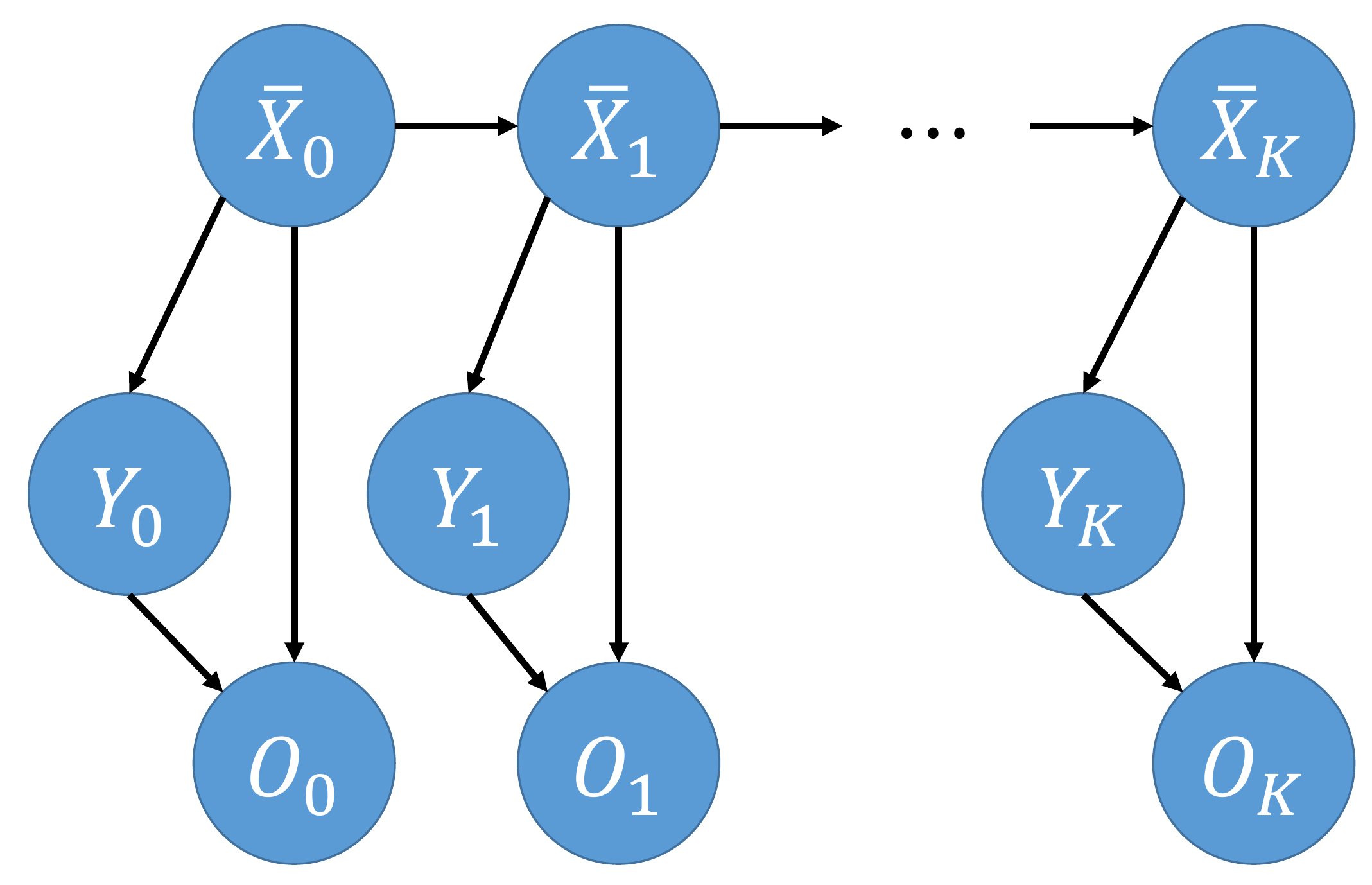}\label{fig:motion_planning}}
	\caption{(a) The GPDM compresses high-dimensional sequential data, $\mathbf{Y}$, into a low-dimensional dynamic system with $\mathbf{X}$, where $\alpha$ and $\beta$ are hyperparameters of the model. (b) The probabilistic model for optimal control; the optimal motion plan can be computed by inferring a posterior trajectory, $\mathbf{X}$, given artificial observations, $\mathbf{O}$, encoding the cost function of control problem. (c, d) The fully-probabilistic model for optimal motion planning with the GPDM (hyperparameters are omitted for visualization sake).}
	\label{fig:grahpical_models}
\end{figure}
This work addresses a motion planning problem of a robot with high-dimensional configuration space.
The objective of the planning problem is to generate a sequence of configurations that achieves a given task and satisfies certain constraints, while being smooth in terms of the robot dynamics.
Rather than solving the planning problem in the original configuration space, we construct a latent variable model with dynamics of much lower number of dimensions and then solved the problem utilizing them.
The latent model with dynamics can be learned from experts' demonstration data or from robot's own experiences; this work particularly considers the cases where the demonstration data is available. %
The learned latent model shown in Fig. \ref{fig:GPDM} represents a probabilistic mapping from the latent space $\mathcal{X}$ to the configuration space $\mathcal{Y}$, and of the stochastic dynamics in the latent space.
A detailed description of latent variable models is given in Section \ref{sec:LVM}.

The model is combined with the probabilistic model for optimal control (shown in Fig. \ref{fig:dual}) that is built using the duality between optimal control and Bayesian estimation.
The intuition behind the duality is that the likelihood of a trajectory for an optimally controlled system is equivalent to the posterior probability when the cost related artificial observation is observed. (More details will be addressed in Section \ref{sec:dual}.)
The combined fully-probabilistic model is shown in Figs. \ref{fig:motion_planning0} and \ref{fig:motion_planning}.
With this, we can convert the original motion planning problem into an inference problem, where the objective is to find the maximum a posteriori (MAP) trajectory given the artificial observation, $O_{1:K}$.
This allows for efficient solution, because every piece of dynamic information is encoded in the low-dimensional latent space.
Descriptions of the combined probabilistic model and proposed solution method are given in Section \ref{sec:MP} and \ref{sec:Alg}, respectively.
Finally, a multiscale acceleration method that increases the algorithm efficiency based on the path integral control is introduced in Section \ref{sec:MS}.
The proposed framework has several advantages;
(a) it does not require knowledge of system dynamics, because this is learned from demonstration data;
(b) rather than exploring the original configuration space \cite{hamalainen2015online}, our framework uses the stochasticity of the latent space for exploration and this provides valuable search space of the motion planning problem;
(c) when computing the most likely trajectory, our framework utilizes a Markov property of the resulting probability model. This provides a more efficient solution method than provided by naive optimization approaches, where the trajectory itself along all the time horizon is considered an (huge) optimization variable~\cite{urtasun2008topologically,wang2008gaussian}.

\section{Constructing a Latent Variable Model} \label{sec:LVM}
\subsection{Gaussian Process Latent Variable Model with Dynamics}
The GP-LVM~\cite{lawrence2005probabilistic} is a generative model that represents a probabilistic mapping from a latent space $\mathcal{X}$ to the observation space $\mathcal{Y}$.
Let $\mathbf{Y}\equiv[\mathbf{y}_1,\mathbf{y}_2,...,\mathbf{y}_N]^T\in\mathbb{R}^{N\times D}$ be the observation matrix, where each row of the matrix represents a single high-dimensional observation of training sequential data, and $\mathbf{X}\equiv[\mathbf{x}_1,\mathbf{x}_2,...,\mathbf{x}_N]^T\in\mathbb{R}^{N\times d}$ be the matrix whose rows represent corresponding latent coordinations of the observations.
In the GP-LVM, the mapping is assumed to be Gaussian process with a covariance function $k_Y(\cdot,\cdot): \mathbb{R}^d\times\mathbb{R}^d\rightarrow\mathbb{R}$, and then the likelihood of the observation data is given by: $p(\mathbf{Y}|\mathbf{X},\beta)=$
\begin{align}
\frac{1}{\sqrt{(2\pi)^{ND}|\mathbf{K}_Y|^D}}\exp\left(-\frac{1}{2}\text{tr}\left(\mathbf{K}_Y^{-1}\mathbf{Y}\mathbf{Y}^T\right)\right), 
\end{align}
where $\mathbf{K}_Y\in\mathbb{R}^{N\times N}$ is a kernel matrix whose components are computed as $(\mathbf{K}_Y)_{ij}=k_Y(\mathbf{x}_i,\mathbf{x}_j)$ with kernel hyperparameters $\beta$.

If the observations are assumed to be obtained from a dynamic system, it would also be possible to construct a probabilistic dynamic model in the latent space~\cite{wang2008gaussian}.
With this model, the likelihood of the latent variable sequence is given by: $p(\mathbf{X}|\alpha)=$
\begin{align}
\frac{p(\mathbf{x}_1)}{\sqrt{(2\pi)^{(N-1)d}|\mathbf{K}_X|^d}}\exp\left(-\frac{1}{2}\text{tr}\left(\mathbf{K}_X^{-1}\mathbf{X}_{2:N}\mathbf{X}_{2:N}^T\right)\right),
\end{align}
where the kernel matrix $\mathbf{K}_X\in\mathbb{R}^{(N-1)\times(N-1)}$ is constructed by $\mathbf{X}_{1:N-1}$ with kernel hyperparameters $\alpha$ and $\mathbf{x}_1$ is assumed to have a Gaussian prior.
Given the prior of hyperparameters, $p(\alpha)$ and $p(\beta)$, a latent variable model with stochastic dynamics is constructed by maximizing the posterior which is proportional to the following joint probability:
\begin{equation}
p(\mathbf{X},\mathbf{Y},\alpha,\beta) = p(\mathbf{Y}|\mathbf{X},\beta)p(\mathbf{X}|\alpha)p(\alpha)p(\beta).
\end{equation}

Note that the GP-LVM is a generative model that ensures smooth mapping from the latent space to the observation space, i.e., the mapping from $\mathcal{X}$ to $\mathcal{Y}$ is continuous and differentiable.
Therefore, the learning process optimizes the model such that two points that are close together in the latent space are mapped to points that are close in the observation space~\cite{lawrence2006local}.
Equivalently, two points that are far apart in the observation space cannot have close latent coordinates, which condition is called \textit{dissimilarity preservation}, and the reverse is only guaranteed when mapping is linear.
Preserving similarity is often considered more important in dimensionality reduction, because observations sparsely lie in high-dimensional space and then similar observations may contain more valuable information between them.
In order to make the GP-LVM have the property of \textit{similarity preservation}, back-constraints were introduced~\cite{lawrence2006local}, which enforces a smooth mapping from the observation space to the latent space.
For example, the kernel based regression model could be used as smooth mapping: $x_{ij} = \sum_{m=1}^{N}a_{jm}k(\mathbf{y}_i,\mathbf{y}_j),$
where closeness in the observation space is measured by the kernel function $k(\cdot,\cdot)$.
Any differentiable mapping (e.g., neural network) could be utilized here; then, the optimization process with respect to the latent coordination $\mathbf{X}$ will simply turn into that with respect to the mapping parameters $\{a_{jm}\}$  using the chain rule.
It is also possible to inject prior knowledge into latent space by restricting a structure of back-constraints~\cite{urtasun2008topologically}.

\subsection{Latent Space Dynamical System}
With the constructed latent variable model, we have the following stochastic dynamics in the latent space:
\begin{equation}
\mathbf{x}_{k+1} = \mu_X(\mathbf{x}_k) + \Sigma_X^{1/2}(\mathbf{x}_k)\mathbf{w}_k, \label{eq:latent_dyn}
\end{equation}
where the mean $\mu_X$ and the variance $\Sigma_X$ are given by the posterior of the Gaussian process:
\begin{align}
&\mu_X(\mathbf{x})=\mathbf{X}_{2:N}\mathbf{K}_X^{-1}\mathbf{k}_X(\mathbf{x}),\nonumber\\
&\Sigma_X(\mathbf{x}) = k_X(\mathbf{x},\mathbf{x})-\mathbf{k}_X(\mathbf{x})^T\mathbf{K}_X^{-1}\mathbf{k}_X(\mathbf{x}),
\end{align}
and $\mathbf{w}_k$ is a $d$-dimensional standard Gaussian random noise.
Here, $\mathbf{k}_X(\mathbf{x})\in\mathbb{R}^{N-1}$ is a vector of which $i$th element represents $k_X(\mathbf{x},\mathbf{x}_i)$.
Moreover, the corresponding pose is also normally distributed:
\begin{equation}
\mathbf{y}_k \sim \mathcal{N}\left(\mu_Y(\mathbf{x}_k),\Sigma_Y(\mathbf{x}_k)\right), \label{eq:observ}
\end{equation}
where the mean $\mu_Y$ and the variance $\Sigma_Y$ are given by
\begin{align}
&\mu_Y(\mathbf{x})=\mathbf{Y}\mathbf{K}_Y^{-1}\mathbf{k}_Y(\mathbf{x}),\nonumber\\
&\Sigma_Y(\mathbf{x}) = k_Y(\mathbf{x},\mathbf{x})-\mathbf{k}_Y(\mathbf{x})^T\mathbf{K}_Y^{-1}\mathbf{k}_Y(\mathbf{x}),
\end{align}
and $\mathbf{k}_Y(\mathbf{x})\in\mathbb{R}^N$ is similarly defined as above.
Combining \eqref{eq:latent_dyn} and \eqref{eq:observ}, the graphical representation of the learned generative model is shown in Fig. \ref{fig:GPDM}.

\section{Optimal Motion Planning in High Dimension via Approximate Inference}
\subsection{Optimal Control via Inference}\label{sec:dual}
Consider a passive and controlled stochastic dynamics,
\begin{align}
\mathbf{x}^{passive}_{k+1} \sim p_k(\cdot|\mathbf{x}_k), \label{eq:passive}\\
\mathbf{x}^{controlled}_{k+1} \sim \pi_k(\cdot|\mathbf{x}_k),
\end{align}
respectively, and the cost rate
\begin{equation}
l_k(\mathbf{x},\pi(\cdot|\mathbf{x})) = q_k(\mathbf{x}) + D_{KL}\left(\pi(\cdot|\mathbf{x})||p(\cdot|\mathbf{x})\right),
\end{equation}
where $q(\mathbf{x})$ is an instantaneous state cost rate that encodes a given task and $D_{KL}(\pi||p)$ is the Kullback–-Leibler (KL) divergence that penalizes a deviation of the controlled dynamics from the passive one.
Then, a stochastic OC problem is formulated with the total cost:
\begin{equation}
J^\pi = \mathbb{E}_{x'\sim\pi(\cdot|x)}\left[q_K(\mathbf{x}_K)+\sum_{k=0}^{K-1}l_k(\mathbf{x}_k,\pi_k(\cdot|\mathbf{x}_k))\right]. \label{eq:cost}
\end{equation}
The above OC problem can be solved by defining the value function,
$$v_k(\mathbf{x}) = \min_\pi\mathbb{E}_{x'\sim\pi(\cdot|x)}\left[q_K(\mathbf{x}_K)+\sum_{\kappa=k}^{K-1}l_\kappa(\mathbf{x}_\kappa,\pi_\kappa(\cdot|\mathbf{x}_\kappa))\right],$$
and solving the Bellman equation on it.
Especially, it is known that a solution of the above OC problem satisfies the linear Bellman equation on the exponentiated value function, called the \textit{desirability function}, $z_k(\mathbf{x})=\exp(-v_k(\mathbf{x}))$ as:
\begin{equation}
z_k(\mathbf{x}) = \begin{cases}
\exp(-q_K(\mathbf{x})), & \text{if }k=K, \\
\exp(-q_k(\mathbf{x}))\mathcal{G}[z_{k+1}(\cdot)](\mathbf{x}), & \text{otherwise}, \\
\end{cases}
\end{equation}
where the linear operator $\mathcal{G}[f(\cdot)](\mathbf{x}) \equiv \int p(\mathbf{x}'|\mathbf{x})f(\mathbf{x}')d\mathbf{x}',$ denotes the average value of the function $f$ at the next time-step~\cite{todorov2009efficient}.
The optimally controlled dynamics is also obtained as:
\begin{equation}
\pi^*_k(\mathbf{x}'|\mathbf{x}) = \frac{p_k(\mathbf{x}'|\mathbf{x})z_{k+1}(\mathbf{x}')}{\mathcal{G}[z_{k+1}(\cdot)](\mathbf{x})}.
\end{equation}
Then, the probability of the trajectory $\mathbf{x}_{1:K}\equiv\{\mathbf{x}_1, \mathbf{x}_2, ..., \mathbf{x}_K\}$ when the state evolves with the optimal transition dynamics $\pi^*$ from the initial state $\mathbf{x}_0$ is given as: $p^*(\mathbf{x}_{1:K}|\mathbf{x}_0)\propto$
\begin{align}
\exp(-q_K(\mathbf{x}_K))\prod_{k=0}^{K-1}\exp(-q_k(\mathbf{x}_k))p_k(\mathbf{x}_{k+1}|\mathbf{x}_k). \label{eq:prop_traj1}
\end{align}
For the detailed derivation, we refer readers to \cite{todorov2009efficient,todorov2011finding} and the references therein.

The OC problem above can be transformed into the Bayesian inference problem.
Suppose we have an artificial binary observation $o_k$ whose emission probability is given by exponential of a negative cost, i.e.,
\begin{equation}
p^o(o_k=1|\mathbf{x}_k) = \exp(-q_k(\mathbf{x}_k)).
\end{equation}
The corresponding graphical model is shown in Fig. \ref{fig:dual}; the transition of the state $\mathbf{x}$ is governed by $p_k$ and it emerges the observation $o$ with $p^o$.
Then, the probability of the trajectory $\mathbf{x}_{1:K}$ given the initial state $\mathbf{x}_0$ and the observation $o_k=1,~\forall k=1,...,K$ is given as: $p(\mathbf{x}_{1:K}|\mathbf{x}_0,o_{1:K} = 1)\propto$
\begin{align}
&\prod_{k=0}^{K-1}p^o(o_{k+1}=1|\mathbf{x}_{k+1})p_k(\mathbf{x}_{k+1}|\mathbf{x}_k)\nonumber\\
&\propto\exp(-q_K(\mathbf{x}_K))\prod_{k=0}^{K-1}\exp(-q_k(\mathbf{x}_k))p_k(\mathbf{x}_{k+1}|\mathbf{x}_k).\label{eq:prop_traj2}
\end{align}
From \eqref{eq:prop_traj1} and \eqref{eq:prop_traj2}, we observe that the OC problem \eqref{eq:passive}--\eqref{eq:cost} is closely related to a Bayesian estimation problem that infers the state trajectory $\mathbf{x}_{1:K}$ when the observation $o_{1:K}=1$ and the initial state $\mathbf{x}_0$ is given (see Fig. \ref{fig:dual})~\cite{rawlik2012stochastic,todorov2011finding}.
The trajectory (or state) distribution induced by the optimal control policy is equivalent to the posterior distribution of the trajectory in the inference problem.
Once the inference problem is formulated, any approximate inference method, such as expectation propagation \cite{toussaint2009robot}, particle belief propagation \cite{hamalainen2015online}, or importance sampling \cite{theodorou2010generalized,ha2016topology}, can be utilized to solve the OC problem efficiently.

\subsection{Motion Planning as MAP Trajectory Estimation}\label{sec:MP}
The framework we propose in this work combines ideas of the latent variable model and the Bayesian interpretation of an OC problem to solve motion planning problems for high-dimensional systems.
The graphical representation of a combined fully-probabilistic model for high-dimensional motion planning problems is shown in Fig. \ref{fig:motion_planning}.
When considering locomotions, the motion data used to learn the latent variable model do not include the global position and orientation of an agent, but only recode their variations, e.g. (angular) velocity, because the learned model needs to encode the dynamics of agent's poses; once the sequence of poses are realized, the global position and orientation can be computed by integrating their variations.
Let $\mathbf{g}\in\mathbb{R}^3$ be the vector of the horizontal position, $(x,~y)$, and heading angle, $\theta$, of the robot and $(\mathbf{y})_v\in\mathbb{R}^3$ denote components corresponding to the forward/lateral velocities and yaw rate of the robot\footnote{$\mathbf{g}$ trivially becomes an empty vector when the problem is not about locomotion.}.
Then, $\mathbf{g}$ also has stochastic dynamics as:
\begin{equation}
\mathbf{g}_{k+1} = \mathbf{g}_k + R_\theta\left(\mu_{Y_v}(\mathbf{x}_k)+\Sigma_{Y_v}^{1/2}(\mathbf{x}_k)\epsilon_k\right), \label{eq:auglatent_dyn}
\end{equation}
where $R_\theta$ is a rotation matrix of the $z$ axis and $\epsilon_k\sim\mathcal{N}(0,I_3)$ follows the standard normal distribution.
Together with the dynamics in the latent space, we define the augmented latent variable as $\bar{\mathbf{x}}\equiv[\mathbf{x}^T, \mathbf{g}^T]^T$, and then the arrows between the above-most nodes in Fig. \ref{fig:motion_planning} represent the temporal structure of the motion planning problem.
Moreover, the latent state at each time step induces robot pose, $\mathbf{y}$ as in \eqref{eq:observ}.
In the planning problem, a cost function, $q_k$, takes robot pose $\mathbf{y}$ as well as its global position and orientation $\mathbf{g}$ as arguments and encodes specifications of a given task, e.g., collision with obstacles or distance from a goal region, etc.
Therefore, $\{\mathbf{y},\mathbf{g}\}$ induces the emission probability of an artificial observation in the proposed probabilistic model in Fig. \ref{fig:motion_planning} as:
\begin{equation}
p^o(o_k=1|\mathbf{y}_k,\mathbf{g}_k) = \exp(-q_k(\mathbf{y}_k,\mathbf{g}_k)).
\end{equation}

In this work, we define the optimal motion plan as the most likely trajectory when the robot's (stochastic) dynamics is optimally controlled.
From the close relationship between \eqref{eq:prop_traj1} and \eqref{eq:prop_traj2}, the problem of finding the most likely trajectory under the optimally controlled dynamics can be converted into the Bayesian estimation problem.
For the transformed estimation problem, the objective is to compute the MAP trajectory $\mathbf{y}^*_{1:K}$ with given initial latent and global coordinates $\bar{\mathbf{x}}_0$ and artificial observations $o_{1:K} = 1$:
\begin{equation}
\{\mathbf{y}_{1:K}^*,\bar{\mathbf{x}}_{1:K}^*\}  = \argmax_{\mathbf{y}_{1:K},\bar{\mathbf{x}}_{1:K}}p(\mathbf{y}_{1:K},\bar{\mathbf{x}}_{1:K}|\bar{\mathbf{x}}_0,o_{1:K} = 1). \label{eq:est_prob}
\end{equation}

\subsection{Dynamic Programming for Computing MAP Trajectory using Particle Filter}\label{sec:Alg}
The estimation problem in \eqref{eq:est_prob} includes two sources of difficulty: long time horizon and continuous space.
The first difficulty can be addressed by exploiting the Markov property of the probability model.
Then, the posterior probability of the trajectory is factorized along the time axis as:
\begin{align}
&\{\mathbf{y}_{1:K}^*,\bar{\mathbf{x}}_{1:K}^*\} \label{eq:est_prob2}\\
&=\argmax_{\mathbf{y}_{1:K},\bar{\mathbf{x}}_{1:K}} \prod_{k=1}^{K}p^o(o_{k}=1|\mathbf{y}_{k},\mathbf{g}_{k})p(\mathbf{y}_{k}|\mathbf{x}_{k})p(\bar{\mathbf{x}}_{k}|\bar{\mathbf{x}}_{k-1}). \nonumber
\end{align}
If the state space is a discrete set, the estimation problem above can be solved via a simple dynamic programming (DP) procedure called the Viterbi algorithm~\cite{koller2009probabilistic}.
However, the observation space $\mathcal{Y}$, as well as the latent space $\mathcal{X}$ in our problem, is continuous; so we need a scheme for approximation of distributions over the continuous space.
In this work, the approximate inference algorithm introduced in \cite{godsill2001maximum} is adopted to address such a difficulty.
The algorithm basically discretizes the continuous state space by the particles obtained from the particle filter (PF) procedure which are expected to span valid regions of the state space due to the resampling procedure.
While the PF procedure builds the approximate discrete state space, the Viterbi algorithm recursively computes the MAP trajectory along the approximated discrete space.

\begin{algorithm}[t]
	\caption{Viterbi algorithm with particle filtering}\label{alg:Viterbi}
	\begin{algorithmic}[1]\small
		\State $\delta_0(1:N) = 0$ \Comment{Initialization}
		\For {$k=1,...,K$} \Comment{Recursion}
		\For {$i=1,...,N_{0}$}
		\State $\mathbf{g}_k^{(i)} \sim \mathcal{N}\left(\mathbf{g}_{k-1}^{(i)}+R_\theta\left(\mathbf{y}_{k-1}^{(i)}\right)_v,R_\theta\Sigma_{Y_v}(\mathbf{x}_{k-1}^{(i)})R_\theta^T\right)$
		\State $\mathbf{x}_k^{(i)} \sim \mathcal{N}\left(\mu_X(\mathbf{x}_{k-1}^{(i)}),\Sigma_X(\mathbf{x}_{k-1}^{(i)})\right)$
		\State $\mathbf{y}_k^{(i)} = \mu_Y(\mathbf{x}_{k}^{(i)})$
		\State $[val, j^*] = \max_j\left[\delta_{t-1}(j)+\log p(\bar{\mathbf{x}}_k^{(i)}|\bar{\mathbf{x}}_{k-1}^{(j)})\right]$
		\State $\delta_k(i)=val-q_k(\mathbf{y}_k^{(i)}, \mathbf{g}_k^{(i)})$
		\State $\psi_k(i) = j^*$
		\State $w_k^{(i)}=w_{k-1}^{(i)}\exp(-q_k(\mathbf{y}_k^{(i)},\mathbf{g}_k^{(i)}))$
		\EndFor
		\State $w_k^{(i)}=w_k^{(i)}/\sum_jw_k^{(j)},~\forall i\in\{1,...,N_{0}\}$
		\State Resample if $\left(\sum_i(w_k^{(i)})^2\right)^{-1} < N_{0}/2$
		\EndFor
		\State $i_T^* = \argmax_i\delta_K(i)$ \Comment{Termination}
		\State $\mathbf{y}_K^* = \mathbf{y}_K^{(i_T^*)}$
		\For {$k=K-1,...,1$} \Comment{Backtracking}
		\State $i_k^* = \psi_{k+1}(i_{k+1}^*)$
		\State $\mathbf{y}_k^* = \mathbf{y}_k^{(i_k^*)}$
		\EndFor
		\State \Return $\mathbf{y}_{1:K}^* = \{\mathbf{y}_1^*, \mathbf{y}_2^*, ... ,\mathbf{y}_K^* \}$
	\end{algorithmic}
\end{algorithm}

The proposed algorithm, shown in Algorithm \ref{alg:Viterbi}, is based on the DP procedure like the Viterbi algorithm, and consists of forward recursion and backtracking.
In the forward recursion, the algorithm constructs the discrete state space by propagating particles of the PF and computes the optimal (partial) trajectories using the DP recursion up to the current time-step.
If the final time is reached, it finds the most-likely final state and the backtracking procedure constructs the optimal trajectory from the chosen state in the backward direction.
In detail, through the forward recursion, the augmented latent variables are propagated as in \eqref{eq:latent_dyn} and \eqref{eq:auglatent_dyn}, and the corresponding pose is also realized (line 4--6).
We restrict the noise in the GP realization \eqref{eq:observ} to $0$, i.e., $\mathbf{y} = \mu_Y(\mathbf{x})$, because this noise only makes resulting motions rougher.
However, this restriction can be easily removed if more complex poses are necessary for the planning.
If we take the log to \eqref{eq:est_prob2}, the equation is simplified further with summation as:
\begin{align}
&\mathbf{y}_{1:K}^* = \{\mu_Y(\mathbf{x}_{k}^*)\}_{k=1,...,K}, \\
&\bar{\mathbf{x}}_{1:K}^* =\argmax_{\bar{\mathbf{x}}_{1:K}} \sum_{k=1}^{K}\log p^o(o_{k}=1|\mu_Y(\mathbf{x}_{k}),\mathbf{g}_{k})p(\bar{\mathbf{x}}_{k}|\bar{\mathbf{x}}_{k-1}) \nonumber
\end{align}
with constraints, $\mathbf{y}_k = \mu_Y(\mathbf{x}_k),~\forall k\in\{1,...,K\}$.
Then the parent nodes that maximize the log-posteriors up to the current state are determined and the log-posteriors of the partial trajectories are computed (line 7--9).
Note that in line 7, $\log p(\bar{\mathbf{x}}_k|\bar{\mathbf{x}}_{k-1})$ has a simple closed form because the dynamics of $\mathbf{x}$ and $\mathbf{g}$ follows the Gaussian distributions as in \eqref{eq:latent_dyn} and \eqref{eq:auglatent_dyn}, respectively.
By computing the weights of particles in line 10, the algorithm can perform resampling procedure which helps the discrete approximation of the space to span valid regions in the original continuous space;
the states having low posterior probabilities (or equivalently, having high costs in the planning problem) up to the current time-step tend not to be re-sampled (line 12--13).
After the forward recursion, the algorithm picks the most likely final state $\mathbf{y}_K^*$ (line 15--16), and then constructs the whole trajectory with backtracking by looking at its ancestry (line 17--20).

\subsection{Multiscale Acceleration with Path Integral Control} \label{sec:MS}
The proposed algorithm approximates the continuous state space at time $k$ by using the PF up to that time.
If only a small number of particles are used in the algorithm, they cannot expand the valid state space as shown in Fig. \ref{fig:MSPI}(a), where all particles failed to reach a goal region.
Consequently, when a problem has a complex cost function (induced by environmental or task complexity), we cannot help but use a large number of particles to span the space effectively.
Having a small number of particles, however, is crucial because the computational complexities of the PF and the DP procedures are $\mathcal{O}(N_0K)$ and $\mathcal{O}({N_0}^2K)$, respectively.
If particles are guided into the better region with useful heuristic, better sample efficiency can be achieved.
This is also closely related to the idea of auxiliary PF~\cite{pitt1999filtering}, where the particles are propagated considering the future measurement.
Trivially, the best sample efficiency can be achieved by propagating particles using the optimal control policy of the original OC problem~\cite{thijssen2015path}, but it is, of course, unrealistic because we do not have the optimal control policy before solving the problem.
\begin{algorithm}[t]
	\caption{PI control with PF at level $l$}\label{alg:PFPI}
	\begin{algorithmic}[1]\small
		\State Input: approximate solution at $(l+1)$ level problem, $\mathbf{u}^{l+1}$.
		\State Create  $\bar{\mathbf{u}}^{l+1}$ by up-sampling $\mathbf{u}^{l+1}$ by a factor of $M_l$.
		\For {$k=1,...,K_l$}
		\For {$i=1,...,N_l$}
		\State $\mathbf{g}_k^{(i)} \sim \mathcal{N}\left(\mathbf{g}_{k-1}^{(i)}+M_lR_\theta\left(\mathbf{y}_{k-1}^{(i)}\right)_v,M_lR_\theta\Sigma_{Y_v}(\mathbf{x}_{k-1}^{(i)})R_\theta^T\right)$
		\State $\mathbf{w}_{k-1}^{(i)} \sim \mathcal{N}(0,I_d)$
		\State $\mathbf{x}_k^{(i)} = \mu_X^l(\mathbf{x}_{k-1}^{(i)},\bar{\mathbf{u}}_k^{l+1})+(\Sigma_X^l(\mathbf{x}_{k-1}^{(i)}))^{1/2}\mathbf{w}_{k-1}^{(i)}$
		\State Append $\{\mathbf{w}_{k-1}^{(i)},\{\mathbf{x}_k^{(i)},\mathbf{g}_k^{(i)}\}\}$ into $\{W^{(i)}_{0:k-1},\bar{X}^{(i)}_{1:k}\}$
		\State $\mathbf{y}_k^{(i)} = \mu_Y(\mathbf{x}_{k}^{(i)})$
		\State $w_k^{(i)}=w_{k-1}^{(i)}\exp(-M_lq_k(\mathbf{y}_k^{(i)},\mathbf{g}_k^{(i)}))$
		\EndFor
		\State $w_k^{(i)}=w_k^{(i)}/\sum_jw_k^{(j)},~\forall i\in\{1,...,N\}$
		\State Resample $\{w^{(i)}_k,W^{(i)}_{1:k},\bar{X}^{(i)}_{1:k}\}$ if $\left(\sum_i(w_k^{(i)})^2\right)^{-1} < N/2$
		\EndFor
		\State $\bar{\mathbf{u}}_{0:K_l-1}^l \gets \{\mathbf{u}_0^l, \mathbf{u}_1^l, ... ,\mathbf{u}_{K_l-1}^l \}$ \Comment{Equation \eqref{eq:PI_con_l}}
		\State \Return $\mathbf{u}_{0:K-1}^l \gets \bar{\mathbf{u}}_{0:K_l-1}^l\otimes\mathbf{1}_{M_l} $
	\end{algorithmic}
\end{algorithm}

In this section, we introduce an efficient multiscale method that guides particles to better regions.
With the discrete time-step $h$, $K=T/h$, and with slight abuse of notation, $q_T(\cdot) = q_K(\cdot)$,  $q_k(\cdot) = q(\cdot,kh)$, the OC problem in \eqref{eq:passive} - \eqref{eq:cost} can be viewed as the discretized version of the following continuous-time OC problem:
\begin{align}
&d\mathbf{x}(t) = F(\mathbf{x}(t))dt + B(\mathbf{x}(t))(\mathbf{u}(t)dt + d\mathbf{w}(t)), \label{eq:conti_sde}\\
&J = \mathbb{E}\left[q_T(\mathbf{x}(T)) + \int_0^{T} q(\mathbf{x}(t),t) + \frac{1}{2}||\mathbf{u}(t)||_2^2dt\right], \label{eq:conti_obj}
\end{align}
because the KL-divergence term in \eqref{eq:cost} can be interpreted as the quadratic control cost, i.e., $D_{KL}(\pi(\cdot|\mathbf{x})||p(\cdot|\mathbf{x})) = \frac{h}{2}||\mathbf{u}||_2^2$\footnote{Note that the KL-divergence between Gaussian distributions $\mathcal{N}(\mu,\Sigma)$ having difference $\mu$ but same $\Sigma$ is given by $\frac{1}{2}(\mu_1-\mu_2)^T\Sigma^{-1}(\mu_1-\mu_2)$ and, in the case of \eqref{eq:conti_sde}, $(\mu_1-\mu_2) = B(\mathbf{x})\mathbf{u}h$ and $\Sigma = B(\mathbf{x})B(\mathbf{x})^Th$, which results in this interpretation.}.
The idea of the proposed method in this section is to sequentially solve approximate discretized problems with different time-steps $M_l\times h$ in the coarse to fine direction, where a solution of a coarser approximate problem is utilized as a priori of a finer approximate problem.
Because the discrete time-length is shorter in the coarser approximate problem, a larger number of particles can be used while maintaining the computational complexity.
We adopt the path integral control method to obtain the optimal control $\mathbf{u}^*$ which also can be solved efficiently within the \textit{``optimal control via inference"} perspective.

Let the original OC problem be level 0 and $M_l\geq 1$ be the number of time steps aggregated in level $l$ for $l=1,...,L$.
That is, $l^\text{th}$ level OC problem is the discretized version of the continuous-time OC problem~\eqref{eq:conti_sde}-\eqref{eq:conti_obj} with time-step $h\times M_l$.
The $l^\text{th}$ level dynamics is guided by the $(l+1)^\text{th}$ level optimal control $\mathbf{u}_k^{l+1}$ and therefore is given by $\mathbf{x}_{k+1}^l = \mu_X^l(\mathbf{x}_k^l, \mathbf{u}_k^{l+1}) + (\Sigma_X^l(\mathbf{x}_k^l))^{1/2}\mathbf{w}_k,~\forall k=1,...,K_l-1$, where
\begin{align}
&\mu^l_X(\mathbf{x},\mathbf{u})=\mathbf{x} + M_l(\mathbf{x}-\mu_X(\mathbf{x})) + M_{l}\Sigma_X^{1/2}(\mathbf{x})\mathbf{u}h,\nonumber\\
&\Sigma^l_X(\mathbf{x}) = M_l\left(k_X(\mathbf{x},\mathbf{x})-\mathbf{k}_X(\mathbf{x})^T\mathbf{K}_X^{-1}\mathbf{k}_X(\mathbf{x})\right),
\end{align}
and $K_l = T/(hM_l) = K/M_l$ is a discrete time-length.
Suppose a set of $N_l$ simulated trajectories and weights $\{w^{(i)}_k,X^{(i)}_{1:k}\}_{i=1,...,N_l}$ is obtained from the PF, which approximates the optimally controlled trajectory distribution.
The path integral control algorithm computes the optimal control solution that minimizes the KL divergence between the controlled trajectory distribution and the optimal one with moment matching~\cite{williams2016aggressive}, and the optimal control is given as:
\begin{align}
\mathbf{u}_k^l \approx \mathbf{u}_k^{l+1} + \frac{1}{hM_lN_l}\sum^{N_l}_{i=1}w_{K_l}^{(i)}\sqrt{M_l}\mathbf{w}_k^{(i)}. \label{eq:PI_con_l}
\end{align}
Such a procedure at level $l$ is summarized in Algorithm \ref{alg:PFPI}.
First, the approximate solution at the upper level $\mathbf{u}^{l+1}$ is up-sampled by a factor of $M_l$ (line 2), where $\mathbf{u}_k^{L+1}=0$ at the coarsest level.
Guided by this, the algorithm propagates particles (line 5--7), realizes poses of robot (line 9), and evaluates particles' weights (line 10, 12).
The algorithm stores the particles' coordinates and the noises that propagated particles in the form of a trajectory, and re-samples them when necessary (line 8 and 13, respectively).
After the filtering procedure, the approximate optimal control is computed as \eqref{eq:PI_con_l}, stretches it back by a factor of $M_l$ and returns it (line 15--16).
With this procedure, the optimal control is sequentially computed from the coarsest level, $l=L$, to the finest level, $l=1$.
Then finally, the finest solution will be utilized to guide the particle propagation of Algorithm \ref{alg:Viterbi}; that is, in the line 6 of Algorithm \ref{alg:Viterbi}, the mean of the particle propagation is changed into $\mathbf{x}_k^{(i)} \sim \mathcal{N}\left(\mu_X(\mathbf{x}_{k-1}^{(i)})+\mathbf{u}^1_{k-1}h,\Sigma_X(\mathbf{x}_{k-1}^{(i)})\right)$.
This guidance is expected to lead that only a small number of particles are enough to expand the valid state space.

\begin{figure}[t]
	\centering
	\subfigure[]{
		\includegraphics*[width=.38\columnwidth]{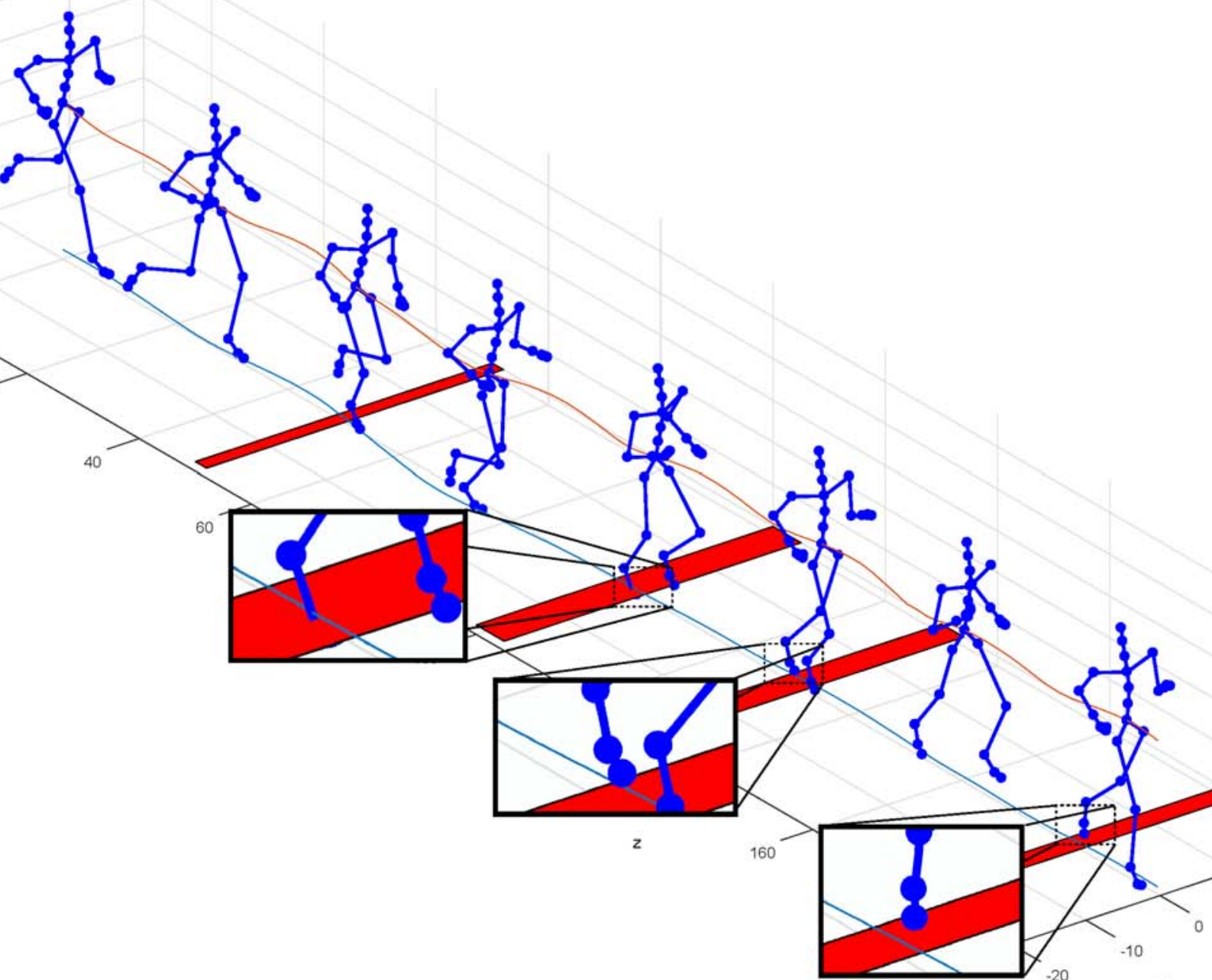}}
	\subfigure[]{
		\includegraphics*[width=.38\columnwidth]{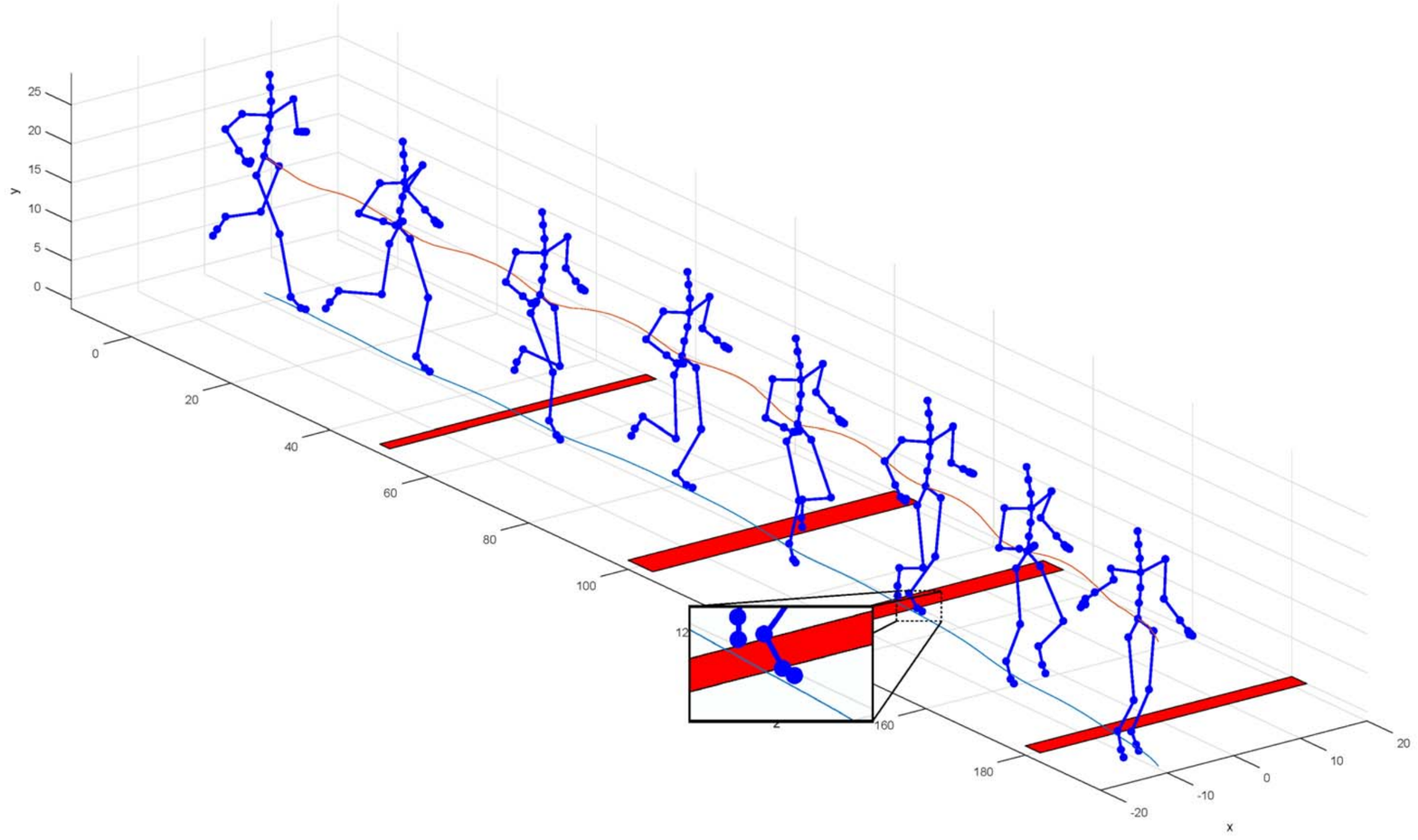}}
	\caption{Resulting trajectories for the experiment 1. The MAP trajectory with the cost function (a) without and (b) with collision checking.}
	\label{fig:jump_env}
\end{figure}
\begin{figure}[t]
	\centering
	\subfigure[]{
		\includegraphics*[width=.38\columnwidth, viewport =110 50 500 310]{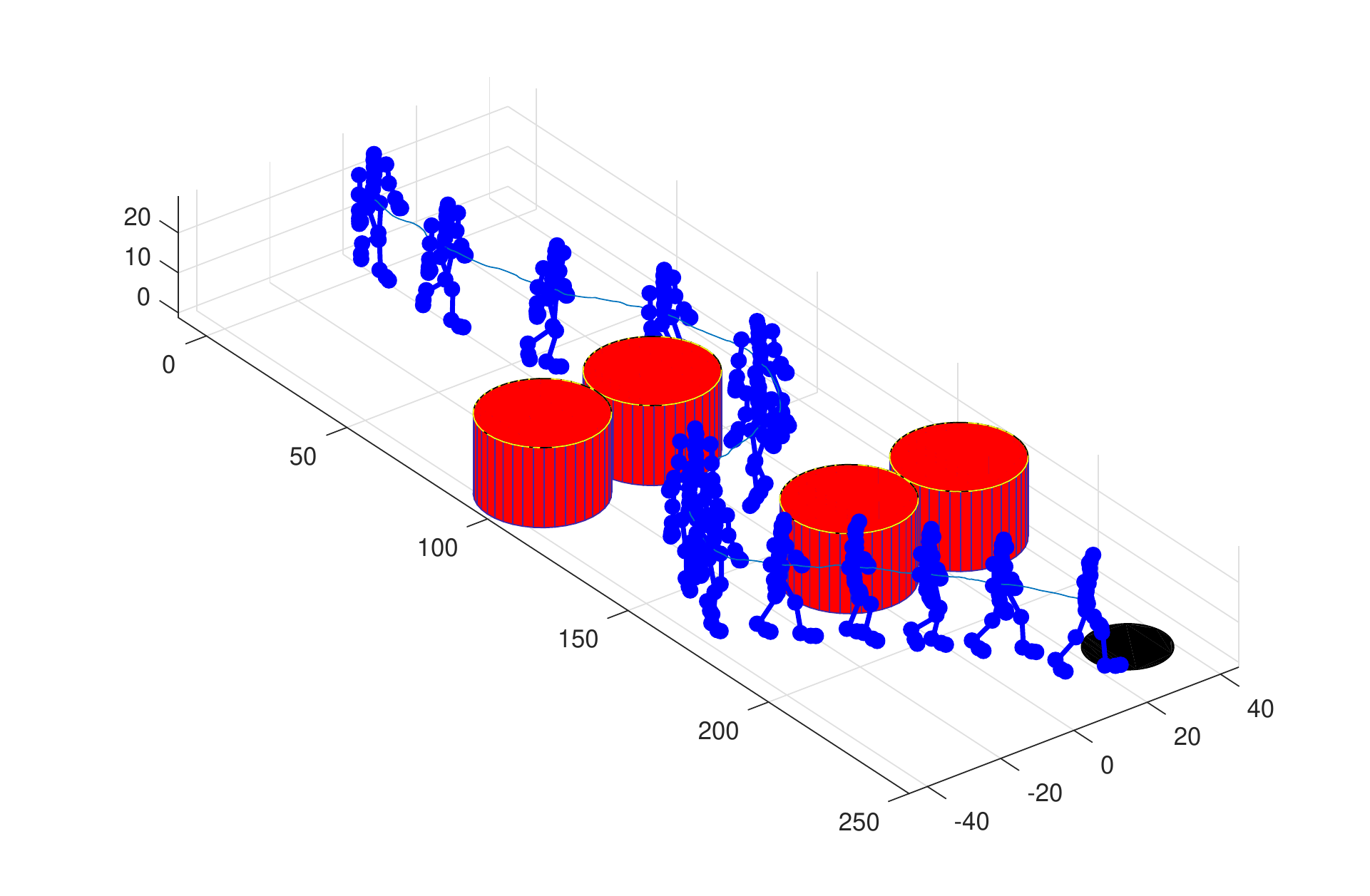}}
	\subfigure[]{
		\includegraphics*[width=.38\columnwidth, viewport =110 50 500 300]{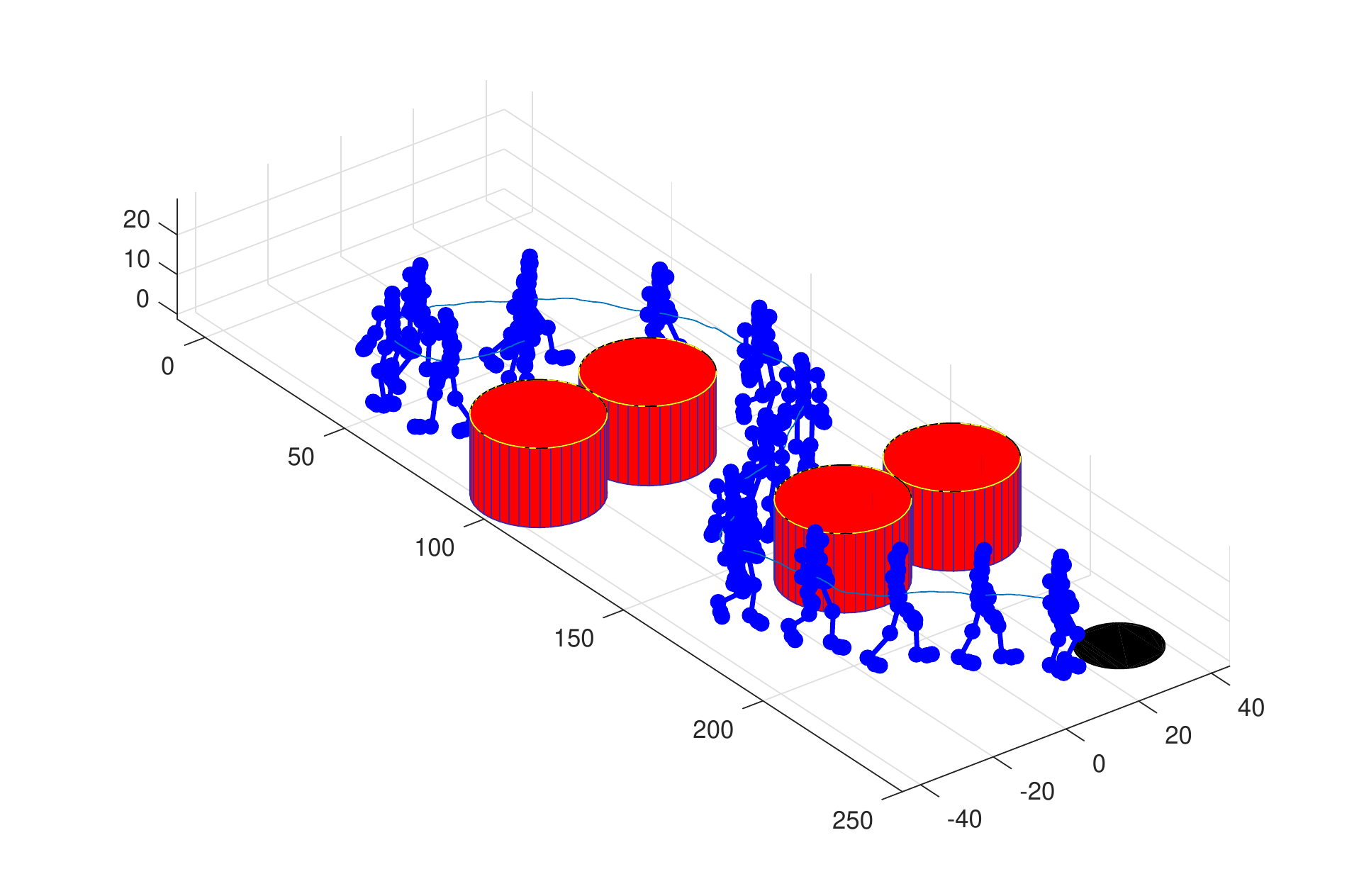}}
	\subfigure[]{
		\includegraphics*[width=.38\columnwidth, viewport =110 50 500 300]{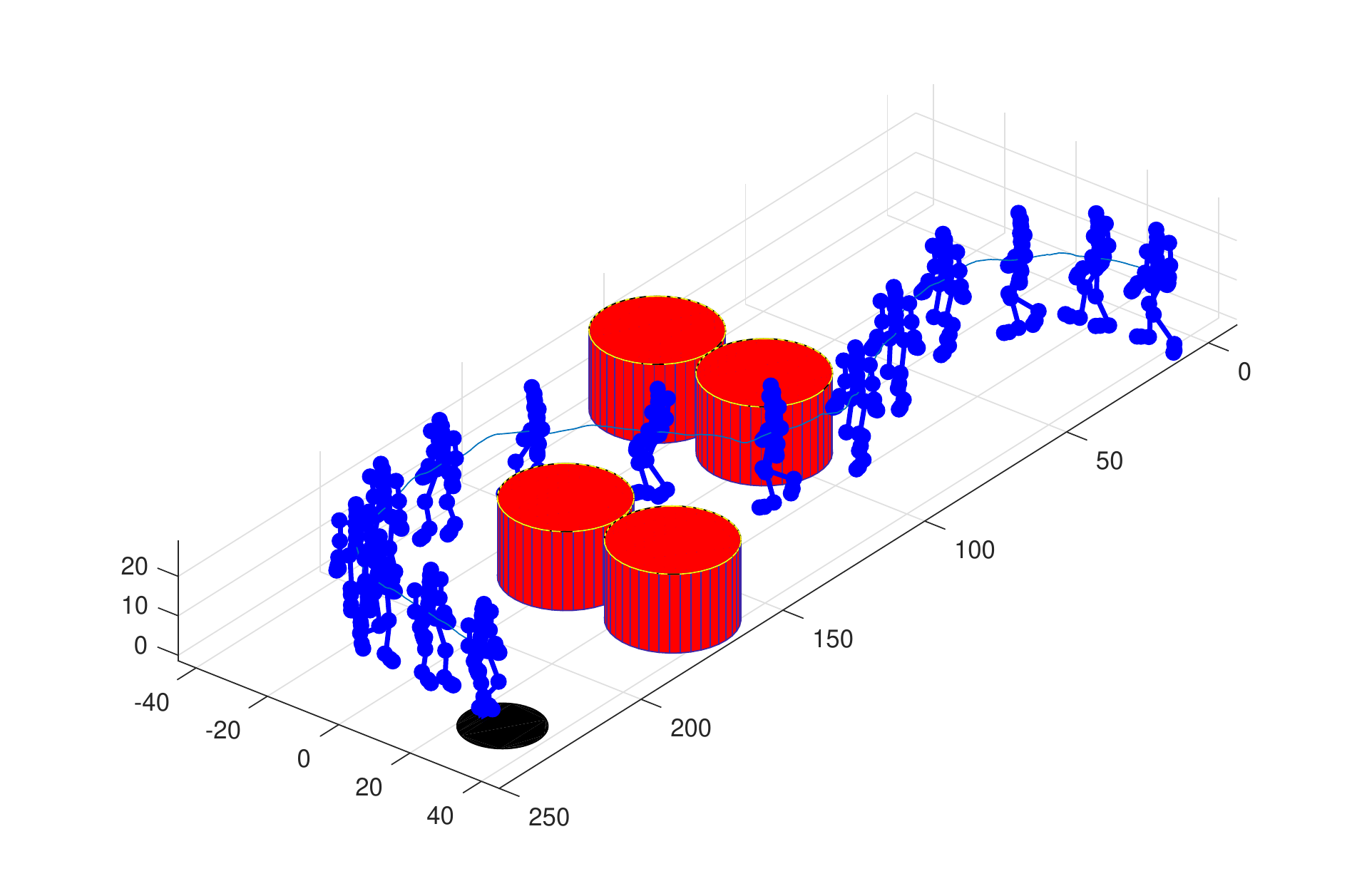}}
	\subfigure[]{
		\includegraphics*[width=.38\columnwidth, viewport =110 60 500 320]{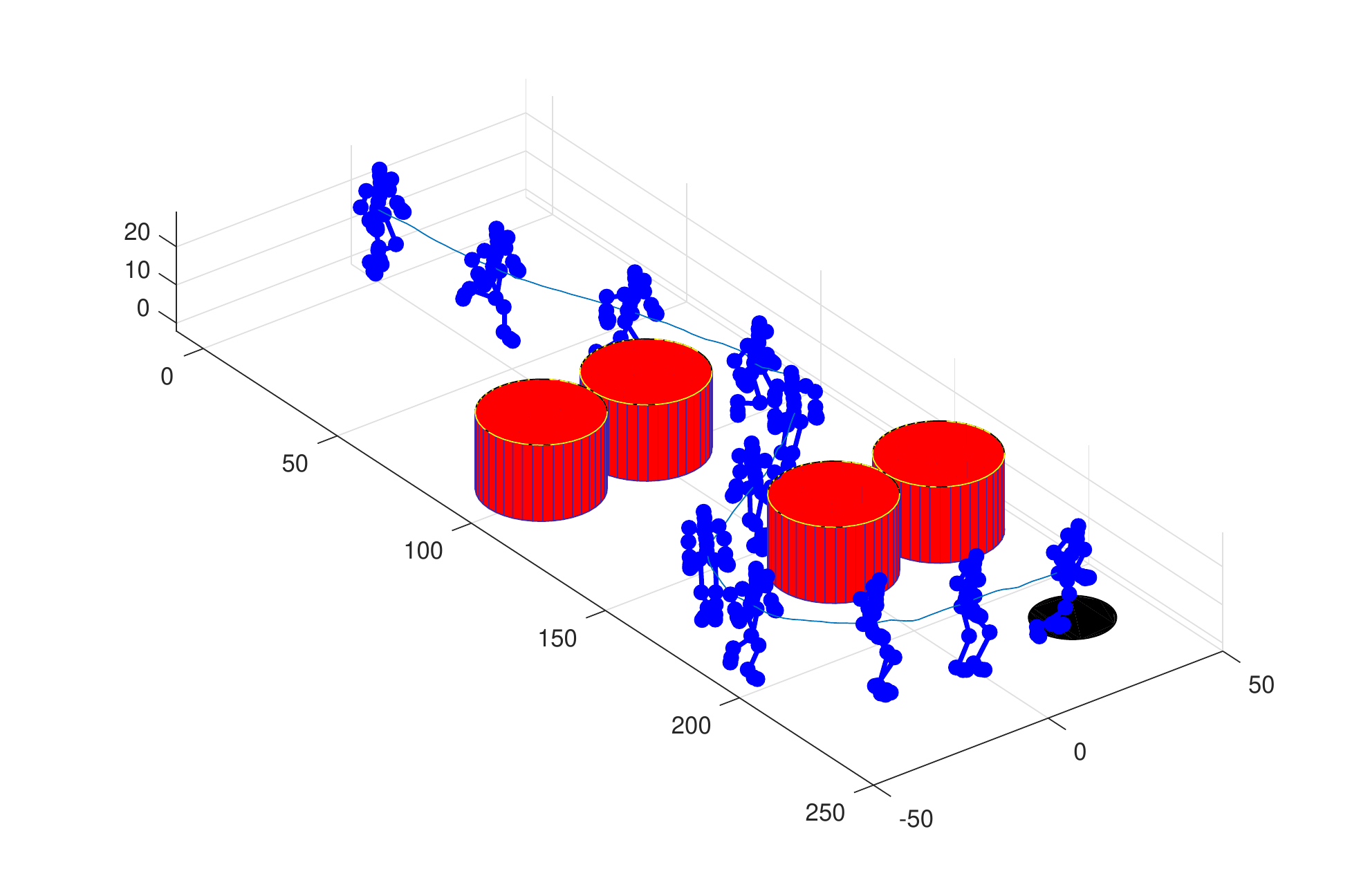}}
	\caption{Resulting trajectories for the experiment 2 from various initial configurations. The goal region is marked by the black circle. The poses are drawn for every (a--c) 30 and (d) 15 steps, i.e., 1 and 0.5 second.}
	\vspace*{-.2in}
	\label{fig:complex_env}
\end{figure}
\begin{figure}[t]
	\centering
	\subfigure[Initial]{
		\includegraphics*[width=.205\columnwidth]{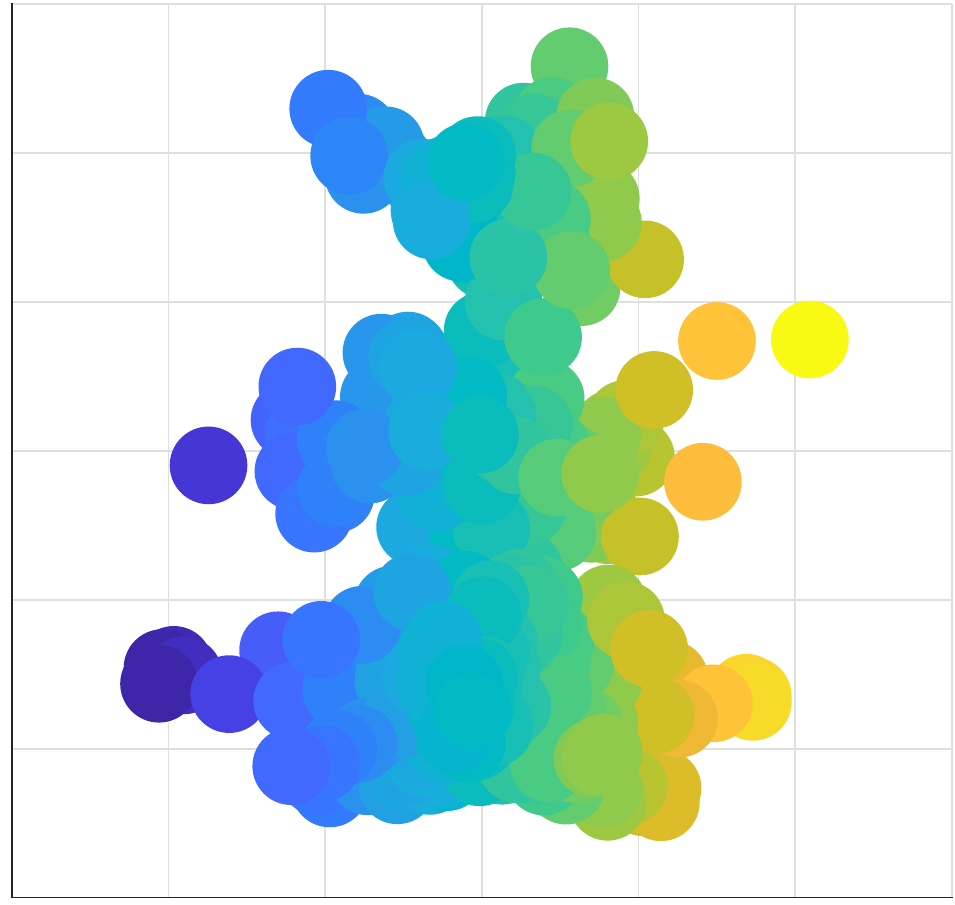}}
	\subfigure[Learned]{
		\includegraphics*[width=.26\columnwidth]{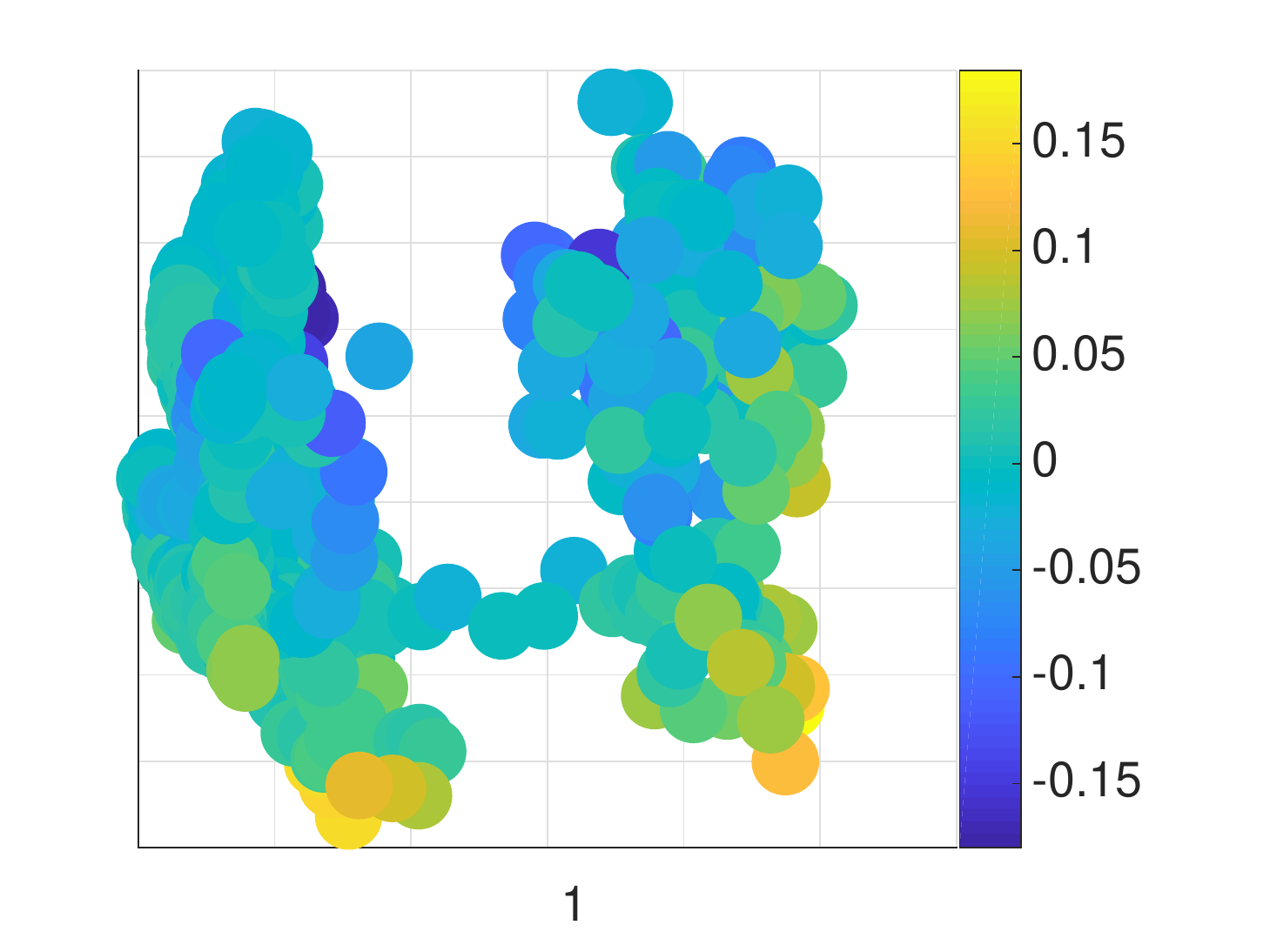}}
	\subfigure[Initial]{
		\includegraphics*[width=.205\columnwidth]{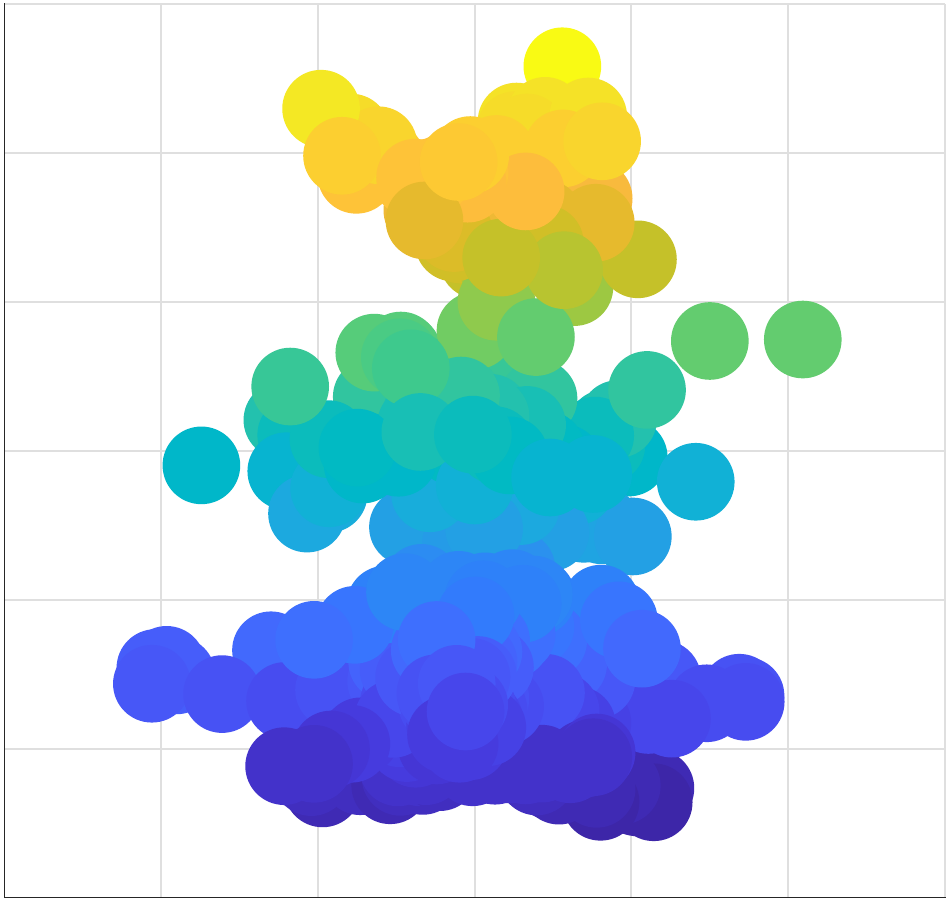}}
	\subfigure[Learned]{
		\includegraphics*[width=.235\columnwidth]{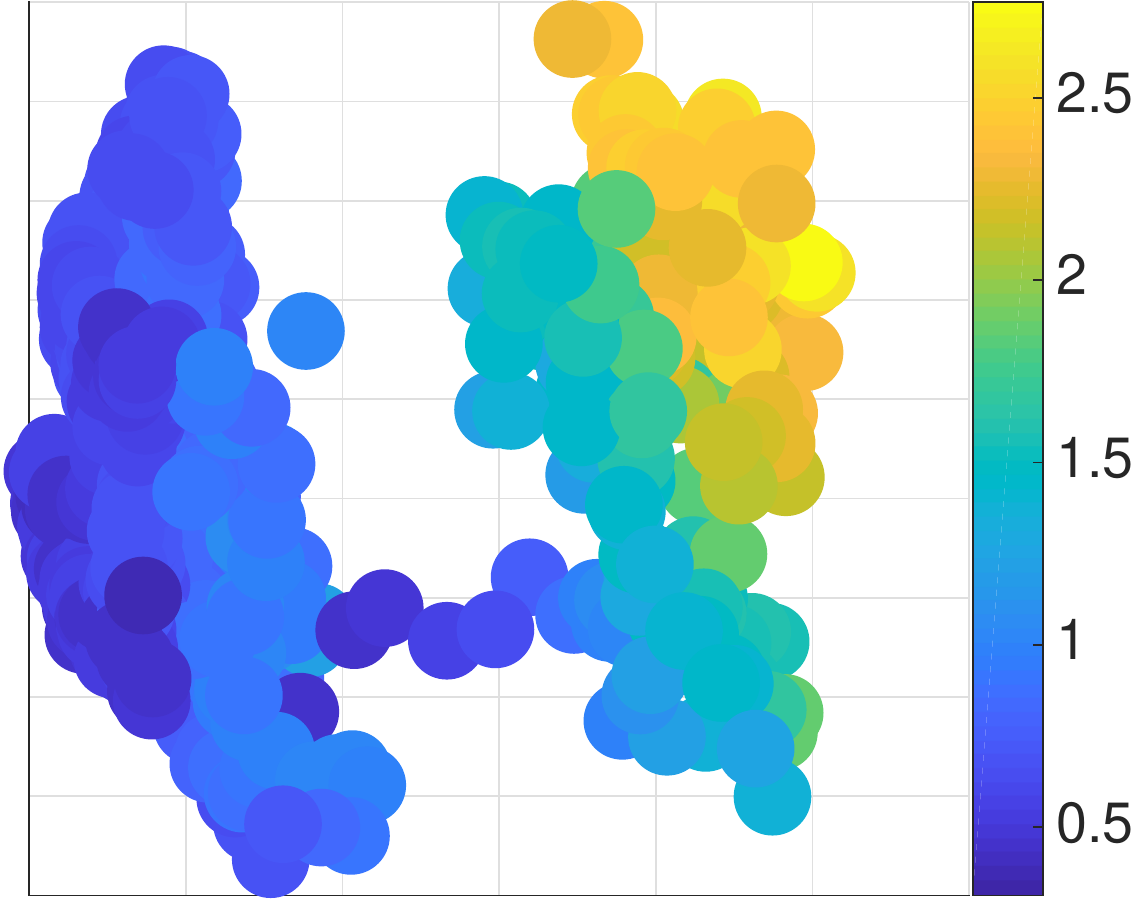}}
	\caption{Representation of various motions in the 4-D latent space projected to the $(x_1, x_2)$ subspace. The points are colored by (a, b) yaw rate and by (c, d) forward velocity of the Mocap data.}
	\vspace*{-.25in}
	\label{fig:latent_sp}
\end{figure}

\section{Example: Humanoid Locomotion Planning}\label{sec:result}
\subsection{Training Data and Hyperparameter Settings}
As an illustrative example, we specifically consider a humanoid robot with $56$-dimensional configuration space.
The Carnegie Mellon University motion capture (CMU mocap) database is used to learn the latent variable model.
The 56-dimensional configurations consist of angles of all joints, roll and pitch angles, vertical position of center of the spine (the root), yaw rate of the root, and horizontal velocity of the root.
The original data were written at 120 Hz, but we down-sampled them into 30 Hz to decrease the size of the observation set while maintaining the quality of motions for capturing the valid temporal information.

For priors of the hyperparameters, we utilized the commonly-used radial basis function (RBF) for $k_X$ and $k_Y$, i.e., $k_X(\mathbf{x},\mathbf{x}') = \alpha_1\exp(-\frac{\alpha_2}{2}||\mathbf{x}-\mathbf{x}'||^2)+\alpha_3^{-1}\delta_{\mathbf{x},\mathbf{x}'},$ and $k_Y(\mathbf{x},\mathbf{x}') = \beta_1\exp(-\frac{\beta_2}{2}||\mathbf{x}-\mathbf{x}'||^2)+\beta_3^{-1}\delta_{\mathbf{x},\mathbf{x}'}$, where $\delta$ is a Kronecker delta function.
As is widely used, we used the uninformative prior on the kernel hyperparameters as $p(\alpha) \propto \prod_i\alpha_i^{-1},~p(\beta) \propto \prod_i\beta_i^{-1}$, which has shown effective regularizations \cite{lawrence2005probabilistic,wang2008gaussian}.
We also used the back-constraints introduced in~\cite{urtasun2008topologically}: because the locomotion has some periodicity, the corresponding latent variable model also does.
First, we extracted the phase of the motion $\phi$ and augmented it to the observation $\mathbf{Y}$.
Then, the back-constraints were used such that the last two latent dimensions had periodic structures as:
$x_{n,d-1} = \sum_{m=1}^N a_m^\text{cos} k_{rbf}(\cos(\phi_n),\cos(\phi_m)) + a_0^\text{cos}\delta_{n,m},$ and $x_{n,d} = \sum_{m=1}^N a_m^\text{sin} k_{rbf}(\sin(\phi_n),\sin(\phi_m)) + a_0^\text{sin}\delta_{n,m}.$
Moreover, the standard RBF back constraints were used in other dimensions.
The GPmat toolbox~\cite{GPmat} is utilized to learn the latent variable models.

\subsection{Results}
For the first numerical experiment, we trained the model using the data sets for walking, fast walking and jogging.
The latent space was set to be $3$-D, where the first and the last two dimensions were initialized as being associated with the forward velocity of the root and the phases, respectively.
Note that the optimization problem of GP-LVMs is non-convex and the learning algorithm is based on the gradient descent method~\cite{lawrence2005probabilistic}, so proper initialization is crucial.
The task for the first experiment was to move forward without stepping on the red lines shown in Fig. \ref{fig:jump_env}.
We set the cost function to create a penalty for the deviation from a desired heading angle, $\theta_d = 0$, a desired $x$ position, $x_d = 0$, forward velocity, $v_d=5m/s$ , and when the foot touched the red lines, i.e., $q(\mathbf{y},\mathbf{g}) = q_\text{obs}(\mathbf{y},\mathbf{g}) + \theta^2 + 0.01|x| + 0.1(v-5)^2. $
The foot positions were computed through the forward kinematics:
$
q_{obs}(\mathbf{y},\mathbf{g}) = \left\{ \begin{array}{ll}
\infty, & \text{if}~FK_{foot}(\mathbf{y},\mathbf{g}) \in D_{red line},\\
0, & \text{otherwise},\end{array} \right.
$
where $FK_{foot}(\cdot)$ is the function of the forward kinematics for the foot joints.
500 particles were used in the Algorithm \ref{alg:Viterbi} with time horizon $K = 90$ (i.e., $T_f = 3$sec.).
In addition, we ignored the randomness in the last 2 dimensions of latent space (which are for phase), i.e., only the learned mean dynamics is used for phase as $\mathbf{x}_{k+1} = \mu_X(\mathbf{x}_k)$, because we found that the noise in those dimensions made the resulting motion too jerky and was not helpful for the planning.
Figs. \ref{fig:jump_env} (a) and (b) show the MAP trajectory without and with collision checking, respectively.
It is shown that the proposed algorithm could find the natural movements that step over the forbidden regions while the motion from the passive dynamics steps on the red lines.
In addition, even though our training data consisted of around 3-4 cycles for each motion, the learned generative model produced natural longer movements continuously.

For the second experiment, we added the datasets for left and right turns to see the trajectory make detours around obstacles.
The latent space was set to be $4$-D here, where the first two dimensions were initialized as corresponding to the yaw rate and forward velocity of the root, while the last two dimensions were for the phases, as in the first example.
The cost function penalized the situations when the robot collided with any obstacle, or left the domain, or reached positions too the distant from the goal region, i.e.,
$
q(\mathbf{y},\mathbf{g}) = q_\text{obs}(\mathbf{y},\mathbf{g}) + q_\text{bnd}(\mathbf{g}) + 10^{-5}q_\text{goal}(\mathbf{g}),
$
where
$q_\text{obs}(\mathbf{y},\mathbf{g})$ and $q_\text{bnd}(\mathbf{g})$ are $\infty$ if $FK(\mathbf{y},\mathbf{g}) \in D_{obstacle}$ and $\mathbf{g}\notin\mathcal{D}$, receptively, and are $0$ otherwise;
$q_\text{goal}(\mathbf{g})$ is the square of the distance to the goal region, computed using the FMT* algorithm~\cite{janson2015fast}.
The domain $\mathcal{D}$ and the goal region were set to be $[-45,45]\times[0, 250]$ and around $(30,230)$, respectively.
1000 particles were used in the Algorithm \ref{alg:Viterbi} with time horizon $K = 450,150$ (i.e., $T_f = 15,5$sec.) for walking and running tasks, respectively.
We ran our algorithm for the cases of various initial positions and orientations and the resulting motion trajectories are shown in Fig. \ref{fig:complex_env}.
It is shown that the proposed algorithm was able to generate smooth and natural motion sequences toward the goal region without collision.
Fig. \ref{fig:latent_sp} shows the learned $4$-D latent space constructed by various motions.
It is observed that, while varying from the initial guess (in Figs. \ref{fig:latent_sp}(a) and (c)), the motions are well-organized in the latent space (in Figs. \ref{fig:latent_sp}(b) and (d)).
We observed that the learned model separated the walking and running motions into two clusters and the transition motion from stand to running was embedded as the small ``bridge" between them.
As a result, the learned generative model hardly produced the transition between walking and running; we expect that more various transition motions are necessary to make more flexible models.
It can be also seen from the supplementary video that the initialized latent variable model cannot generate proper natural motions.
We would refer the readers to the supplementary video for more visualization.

\begin{figure}[t]
	\centering	
	\subfigure[]{
		\includegraphics*[width=.15\columnwidth, viewport =30 50 400 600]{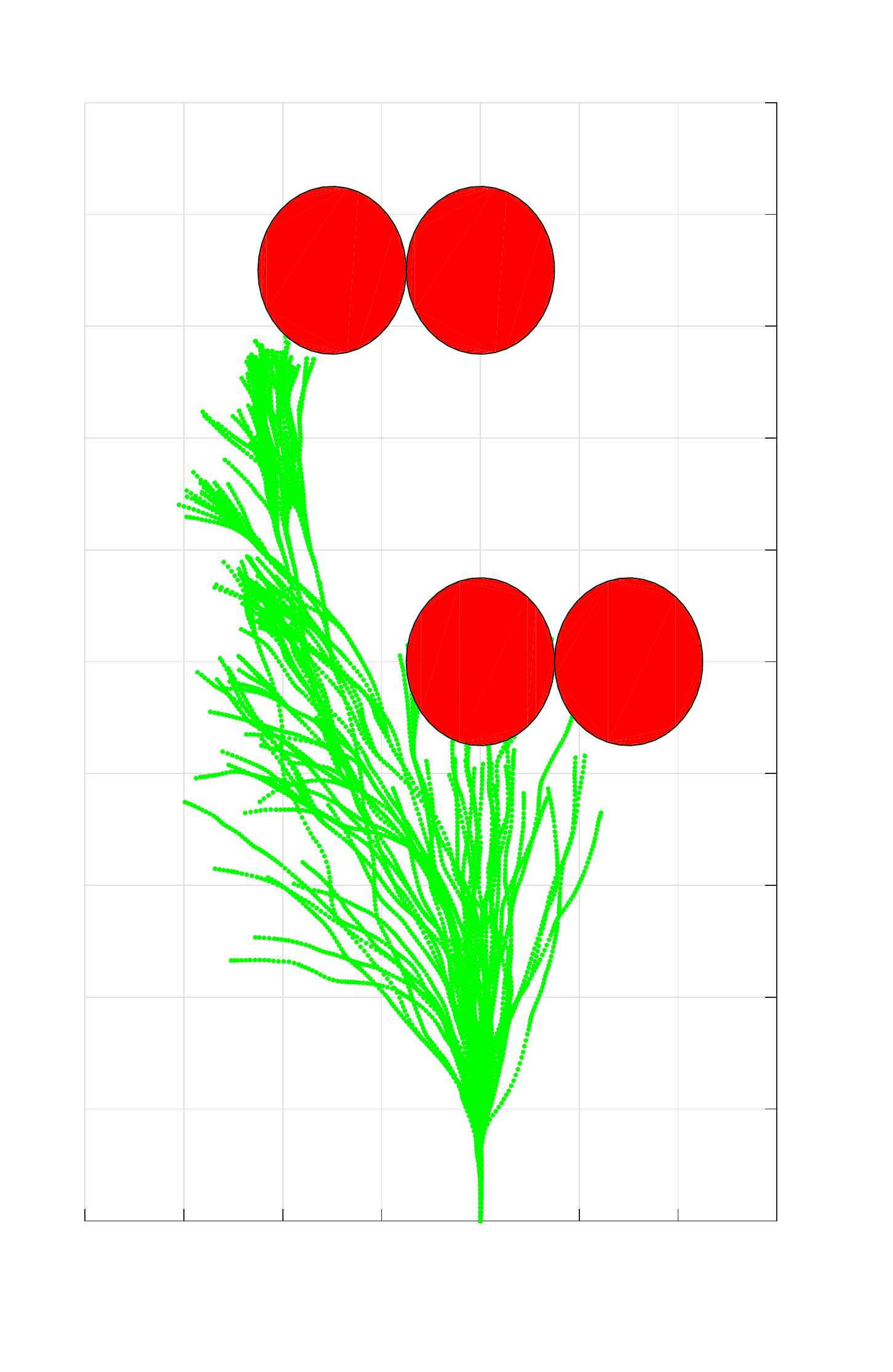}}
	\subfigure[]{
		\includegraphics*[width=.15\columnwidth, viewport =30 50 400 600]{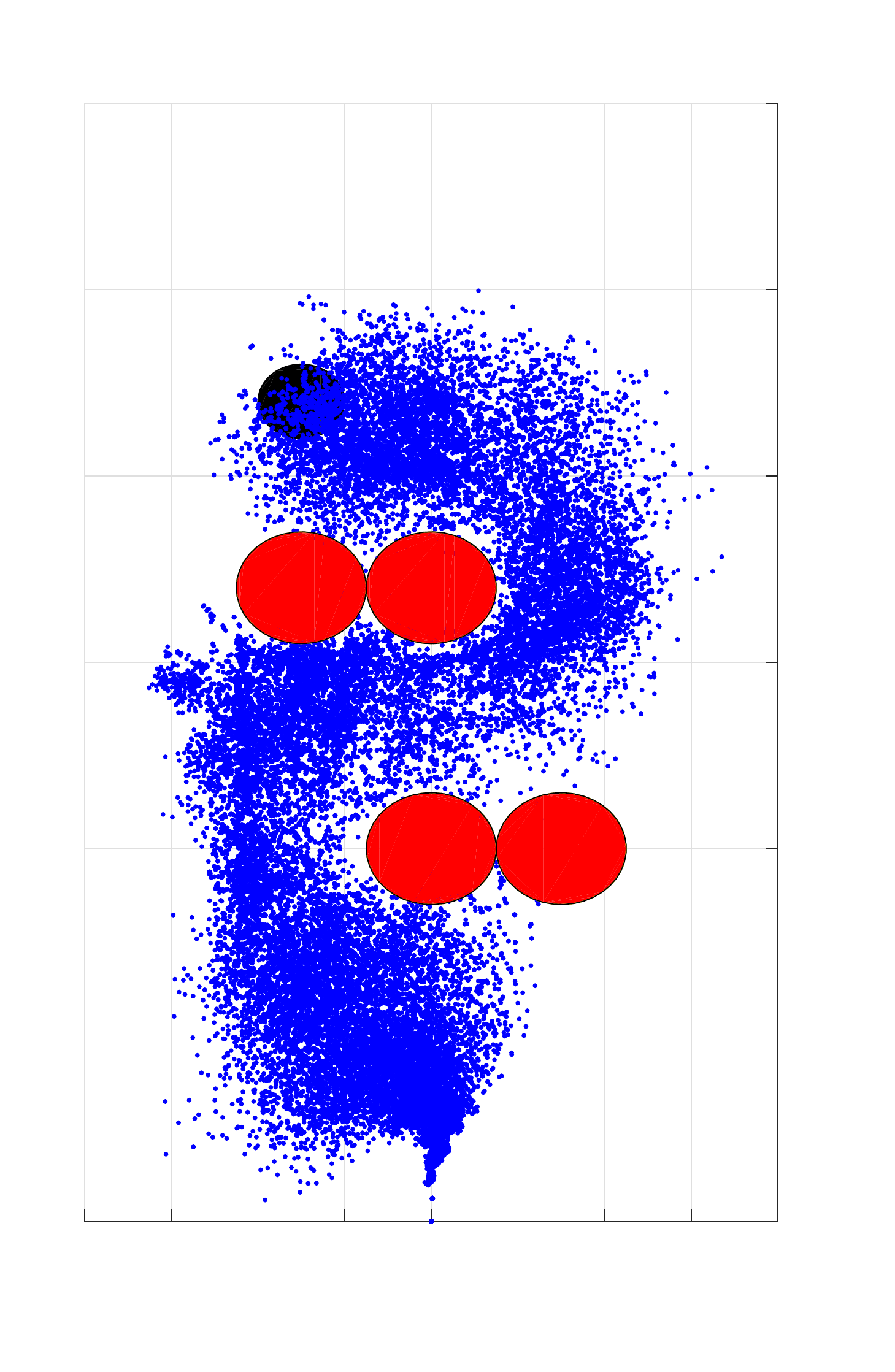}}
	\subfigure[]{
		\includegraphics*[width=.15\columnwidth, viewport =30 50 400 600]{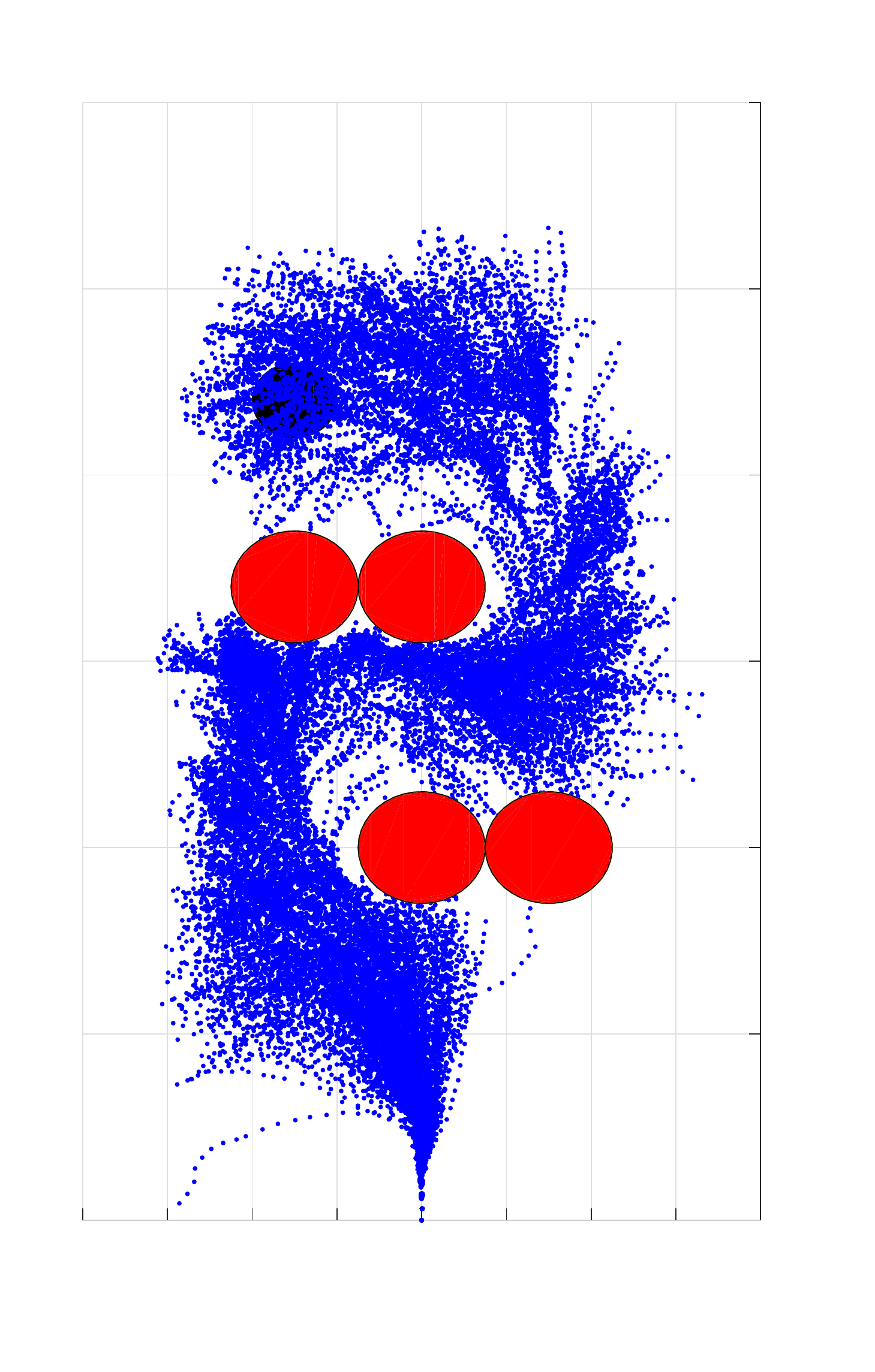}}
	\subfigure[]{
		\includegraphics*[width=.15\columnwidth, viewport =30 50 400 600]{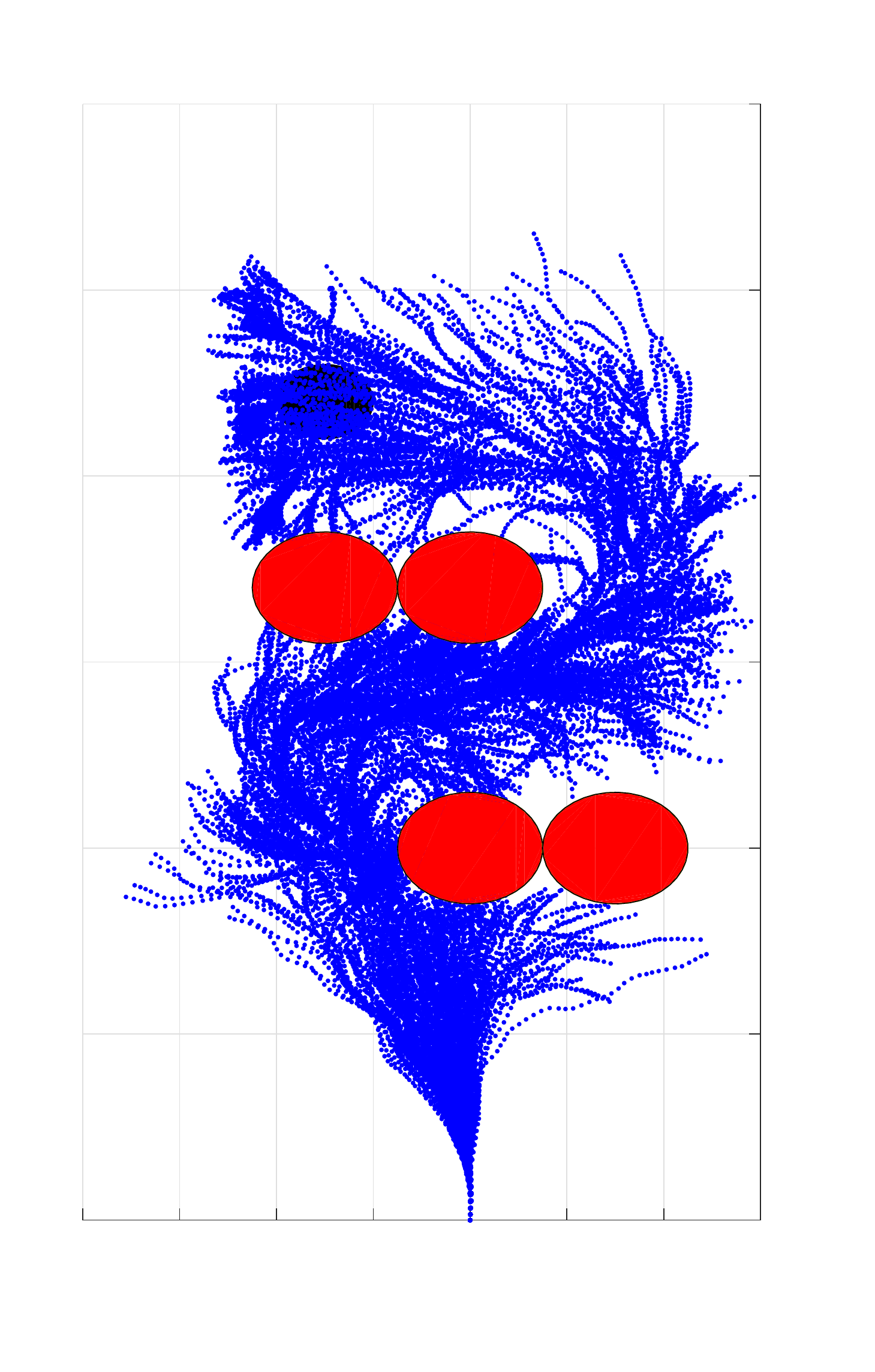}}
	\subfigure[]{
		\includegraphics*[width=.155\columnwidth, viewport =30 50 400 600]{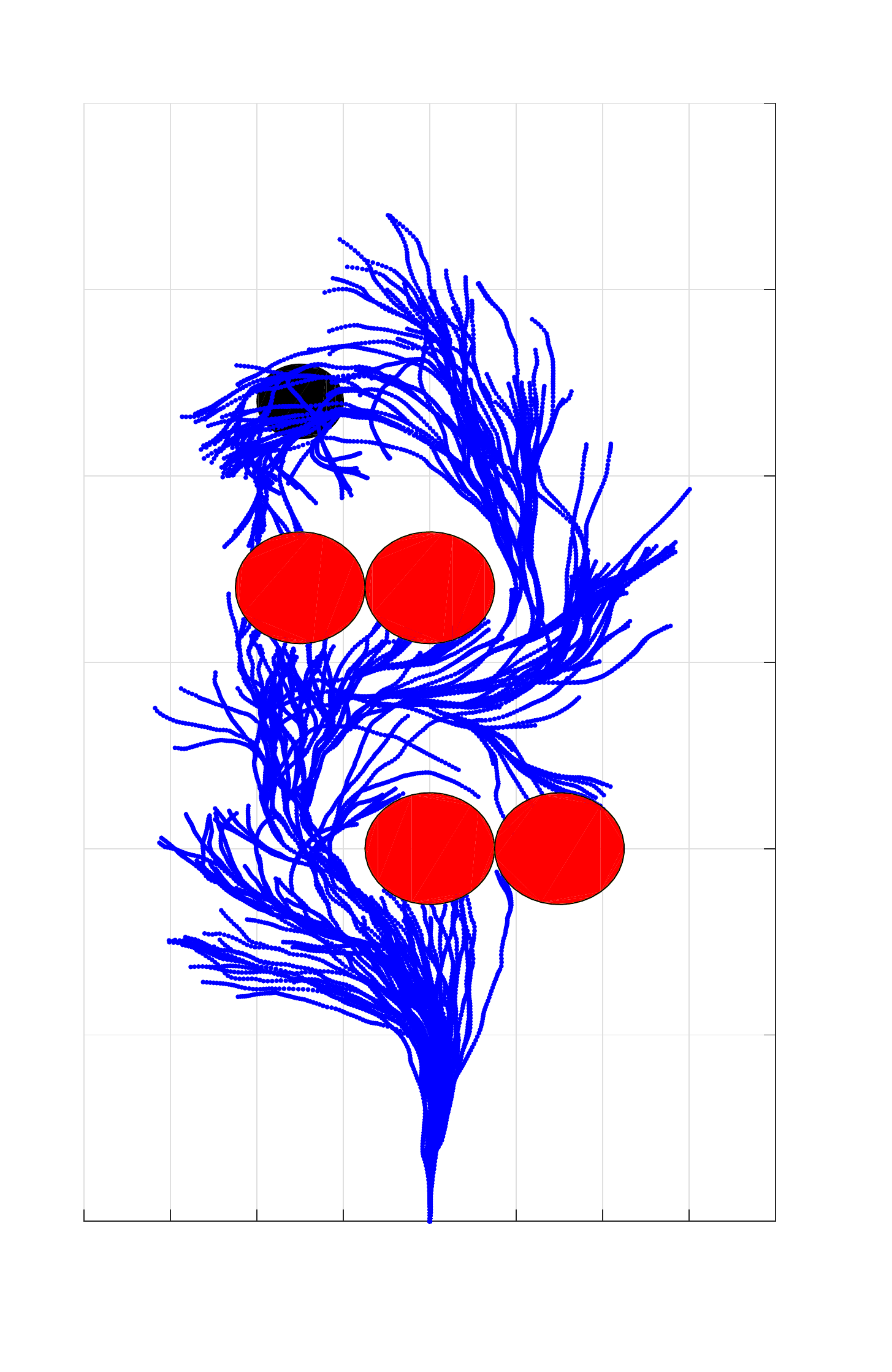}}
	\caption{The particles from (a) the naive PF approach with 50 particles and from the recursive multiscale procedure with (b) $M_3 = 8,~N_3=800$ (c) $M_2 = 4,~N_2=400$ (d) $M_1 = 2,~N_1=200$ (e) $M_0 = 1,~N_0=50$.}
	\vspace*{-.05in}
	\label{fig:MSPI}
\end{figure}
\begin{table}[t]
	\caption{Computational complexity and success rate}
	\vspace*{-.15in}
	\label{tbl:MSPI}
	\begin{center}
		\begin{tabular}{l ccc ccc}
			\hline
			\hline
			Case & PF & DP & CPU (sec) & Env. 1 & Env. 2 & Env. 3\\
			\hline
			$N_0=50$ & $1$ & $1$ & 13.25 & 100\% & 23\% & 9\% \\
			$N_0=500$ & $10$ & $100$ & 140.16 & 100\% & 82\% & 77\% \\
			$N_0=1000$ & $20$ & $400$ & 262.46 & 100\% & 98\% & 95\% \\
			Multi-Scale & $7$ & $1$ & 87.17 & 100\% & 85\% & 85\% \\
			\hline
			\hline
		\end{tabular}
	\vspace*{-.25in}
	\end{center}
\end{table}
Finally, we tested multiscale acceleration method.
The multiscale parameters were set as $N_l = 200,~400,~800$ and $M_l = 2,~4,~8$ and $N_0=50$ particles are used in the original scale.
These multiscale parameters should be chosen carefully, because the approximate solution \eqref{eq:PI_con_l} may fail to guide the finer level dynamics properly if a gap between the scales is too large.
Figs. \ref{fig:MSPI}(b)-(e) depict the particles from the proposed recursive procedure.
It is observed that as the approximation level decreases, smaller number of particles are well-guided to the goal region; the larger number of particles successfully expanded the state space at the higher level, and then only 50 particles succeeded to be propagated toward the goal at the original scale.
We also compared the performances of Algorithm 1 with various numbers of particles ($N_0=50,~500,~1000$) in various environments having different obstacles: the environment 3 is same as that of Fig. \ref{fig:complex_env}(a), the environment 2 only had 2 obstacles close to the goal region, and the environment 1 had no obstacles; the initial state and the goal region were same as those of Fig. \ref{fig:complex_env}(a).
The comparison result is shown in Table~\ref{tbl:MSPI}.
The second, third and fourth columns denote the relative computational complexities of PF, i.e. $\mathcal{O}(N_0K)$, DP, i.e. $\mathcal{O}({N_0}^2K)$, and computation time for whole planning procedures, respectively.
Note that, for the algorithm with the multiscale acceleration, the computational complexity of DP is same to the case of $N_0=50$ because the same number of particles are used, and the computational complexity of PF is $7$ times larger than the case of $N_0=50$ because those are twice at each level $N_l/M_l = 100,~\forall l=1,2,3$ and same at the original level.
We found that most computational budgets were spent by GP realizations (around $50\%$) and forward kinematics for collision checking (around $30\%$), which is proportional to the number of data $N$, particles $N_0$, and the planning horizon $K$.
Therefore, the planing algorithm can be accelerated further by using ideas of sparse GP-LVMs~\cite{lawrence2005probabilistic} and more efficient forward kinematics and collision checking methods; we leave it as future works.
The last three columns show the rate of each algorithm that at least one of particles achieved the goal region.
Table~\ref{tbl:MSPI} shows that though a larger number of particles need to be used as the environment becomes complex, which results in huge computational complexity, the proposed multiscale method can guarantee the performance while maintaining the computational complexity properly.

\section{Conclusions}
In this work, we proposed an efficient framework for generating the motion trajectory of a robot with high degrees of freedom.
The framework included a probabilistic generative model for the motion sequence from demonstration data using the GP-LVM method.
The constructed low-dimensional model was then combined with the probabilistic model of the OC problem.
Finally, we proposed an efficient approximate MAP trajectory estimation algorithm modified to utilize the dynamic programming procedure and the multiscale path integral control method to increase the sample efficiency.


\addtolength{\textheight}{-12cm}   






\bibliographystyle{IEEEtran}
\bibliography{RAL18}

\end{document}